\documentclass[preprint,review,12pt]{elsarticle}

\usepackage[table]{xcolor}
\usepackage{longtable}
\usepackage{amssymb}
\usepackage{siunitx}
\usepackage{amsmath}
\usepackage{arydshln}
\usepackage{multirow}
\usepackage{booktabs}
\usepackage{hyperref}
\usepackage{graphicx} 
\usepackage{nameref}
\usepackage{wrapfig}
\usepackage{tabularx}
\usepackage[ruled,vlined,linesnumbered]{algorithm2e}
\usepackage{tcolorbox} 
\usepackage{cleveref}
\usepackage{setspace}
\usepackage{placeins} 
\usepackage{pgfplots}
\pgfplotsset{compat=1.18}
\usepackage{enumitem}
\usepackage{geometry}  
\SetCommentSty{mycommfont}

\newcommand{\dr}{\text{InitialRating}}
\newcommand{\rg}{\text{rating}} 
\newcommand{\ra}{\text{rating}[c]} 
\newcommand{\rpl}{\text{rating}^l[c]} 
\newcommand{\rmi}{\text{rating}^r[c]} 
\newcommand{\rb}{\text{rating}^{l/r}[c]}
\newcommand{\lr}{\text{localRating}}
\newcommand{\ras}{\text{rating}_s[c]}
\newcommand{\rpls}{\text{rating}_s^l[c]}
\newcommand{\rmis}{\text{rating}_s^r[c]}
\newcommand{\rbs}{\text{rating}_s^{l/r}[c]}
\newcommand{\lrs}{\text{localRating}_s}
\newcommand{\rbx}{\text{origRating}^\pm[c]}
\newcommand{\av}{\text{avgRating}}
\newcommand{\avd}{\text{avgRating}[d]}
\newcommand{\cle}{n^l[c]}
\newcommand{\cri}{n^r[c]}
\newcommand{\clr}{n^{l/r}[c]}

\newcommand{\varm}{$x$}
\newcommand{\var}{x}
\newcommand{\optsol}{OptalCP}
\newcommand{\cpsol}{CP Optimizer}
\newcommand{\pA}{\emph{Plan~A}}
\newcommand{\pB}{\emph{Plan~B}}
\newcommand{\FF}{\texttt{Fail-First}}
\newcommand{\C}{\mathcal{C}}
\newcommand{\Pe}{\mathcal{P}}
\newcommand{\V}{\mathcal{V}}

\newcommand{\eg}{$\epsilon$-greedy}
\newcommand{\bg}{$B$-greedy}
\newcommand{\ug}{$U$-greedy}
\newcommand{\tg}{$T$-greedy}
\newcommand{\egL}{$\epsilon_{3\%}$-greedy}

\newcommand{\egHR}{$\epsilon_{10\%R}$-greedy}

\newcommand{\bgL}{$B_{3\%}$-greedy}

\newcommand{\bgHR}{$B_{10\%R}$-greedy}

\newcommand{\tgL}{$T_{3\%}$-greedy}
\newcommand{\tgLR}{$T_{3\%R}$-greedy}

\newcommand{\jsspMab}{1.5} 
\newcommand{\rcpspMab}{2.2} 
\newcommand{\jsspTuning}{1.1} 
\newcommand{\rcpspTuning}{1.2} 

\newcommand{\jsspToOldFDS}{1.7} 
\newcommand{\rcpspToOldFDS}{2.5} 
\newcommand{\jsspToCPOpt}{3.5} 
\newcommand{\rcpspToCPOpt}{2.1} 

\newcommand{\repeatedSavings}{\SI{50}{\percent}}
\newcommand{\repeatedLastSavingsJSSP}{\SI{30}{\percent}}
\newcommand{\repeatedLastSavingsRCPSP}{\SI{38}{\percent}}

\newcommand{\jsspBetter}{78}  
\newcommand{\jsspAll}{84}  
\newcommand{\rcpspBetter}{226}  
\newcommand{\rcpspAll}{393}  
\newcommand{\jsspCompleteHardSet}{161}
\newcommand{\rcpspCompleteHardSet}{200}
\newcommand{\jsspTrainingSet}{81}
\newcommand{\rcpspTrainingSet}{100}
\newcommand{\jsspTestingSet}{80}
\newcommand{\rcpspTestingSet}{100}

\SetKwData{hp}{Choices}               
\SetKwData{st}{BranchStack}           
\SetKwData{bc}{Choice}                
\SetKwData{br}{Branch}                
\SetKwData{lbr}{LeftBranch}           
\SetKwData{tb}{TrueBranch}            
\SetKwData{rbr}{RightBranch}          
\SetKwData{sol}{Solution}             
\SetKwData{ub}{UB}                    
\SetKwData{fc}{FailCount}             
\SetKwData{fl}{FailLimit}             
\SetKwData{rf}{RestartFactor}         
\SetKwFunction{pp}{Pop()}             
\SetKwFunction{ps}{Push}              

\SetKwRepeat{Do}{do}{while}

\SetKwFunction{GIH}{GenerateInitialChoices}     
\SetKwFunction{GMC}{GenerateAdditionalChoices}        
\SetKwFunction{SC}{SelectChoice}                
\SetKwFunction{ID}{IsDecided}                   
\SetKwFunction{GC}{GetChoice}                   
\SetKwFunction{UBR}{UpdateBranchRating}         
\SetKwFunction{CIL}{CurrentIsLeft}                      
\SetKwFunction{PC}{Propagate}                   
\SetKwFunction{RB}{RotateBranches}                   
\SetKwFunction{FFA}{FindFeasibleAssignment}     
\SetKwFunction{LO}{LimitObjective}              
\SetKwFunction{GNGC}{GenerateNoGoodConstraints}  

\journal{Computers \& Industrial Engineering}

\begin{document}

\begin{frontmatter}

\title{Reinforcement Learning for Search Tree Size Minimization in Constraint Programming: New Results on Scheduling Benchmarks}

\author[inst1,inst3]{Vilém Heinz\texorpdfstring{\corref{cor1}}{}}
\ead{vilem.heinz@cvut.cz}
\cortext[cor1]{Corresponding author}

\author[inst2]{Petr Vilím}

\author[inst3]{Zdeněk Hanzálek}

\affiliation[inst1]{
    organization={Faculty of Electrical Engineering, Czech Technical University in Prague},
    addressline={Technicka 1902/2, Prague 6}, 
    city={Prague},
    postcode={166 27}, 
    country={Czech Republic}
}

\affiliation[inst2]{
    organization={ScheduleOpt},
    addressline={Pod zamkem 103}, 
    city={Novy Knin},
    postcode={262 03}, 
    country={Czech Republic}
}

\affiliation[inst3]{
    organization={Czech Institute of Informatics, Robotics and Cybernetics, Czech Technical University in Prague},
    addressline={Jugoslavskych partyzanu 1580/3, Prague 6}, 
    city={Prague},
    postcode={160 00}, 
    country={Czech Republic}
}

\begin{abstract}
Failure-Directed Search (FDS) is a significant complete generic search algorithm used in Constraint Programming (CP) to efficiently explore the search space, proven particularly effective on scheduling problems.
This paper analyzes FDS's properties, showing that minimizing the size of its search tree guided by ranked branching decisions is closely related to the Multi-armed bandit (MAB) problem. 
Building on this insight, MAB reinforcement learning algorithms are applied to FDS, extended with problem-specific refinements and parameter tuning, and evaluated on the two most fundamental scheduling problems, the Job Shop Scheduling Problem (JSSP) and Resource-Constrained Project Scheduling Problem (RCPSP).
The resulting enhanced FDS, using the best extended MAB algorithm and configuration, performs \jsspToOldFDS\ times faster on the JSSP and \rcpspToOldFDS\ times faster on the RCPSP benchmarks compared to the original implementation in a new solver called \optsol, while also being \jsspToCPOpt\ times faster on the JSSP and \rcpspToCPOpt\ times faster on the RCPSP benchmarks than the current state-of-the-art FDS algorithm in IBM \cpsol\ 22.1.
Furthermore, using only a \SI{900}{\second} time limit per instance, the enhanced FDS improved the existing state-of-the-art lower bounds of \jsspBetter\ of \jsspAll\ JSSP and \rcpspBetter\ of \rcpspAll\ RCPSP standard open benchmark instances while also completely closing a few of them.
\end{abstract}


\begin{keyword}
Constraint Programming \sep Reinforcement Learning \sep Discrete Optimization \sep Scheduling \sep Tree Search \sep Heuristics 


\MSC[2020] 90-08 \sep 90B35 \sep 90C59 \sep 90C99 \sep 68T20 \sep 90C27

\end{keyword}

\end{frontmatter}


\section{Introduction}
\label{sec:intro}
Constraint Programming (CP) is a declarative programming paradigm similar to Integer Linear Programming (ILP).
It is used for various combinatorial problems, but it particularly excels in scheduling problems.
Its application in this area often yields state-of-the-art results for both well-known general problems like \cite{LIESS, HE20123331, icores25}, as well as for more specific problems with many additional constraints like \cite{ABREU2022108128, ROHANINEJAD2023108958, HEINZ2022108586}, usually outperforming ILP \cite{LUNARDI2020105020, FATEMIANARAKI2023102770}.

This paper discusses Failure-Directed Search (FDS), a complete generic search algorithm designed to explore the search space of CP problems systematically and efficiently.
FDS was introduced in \cite{vilim2015failure} as an algorithm to prove optimality / infeasibility (and lower bounds\footnote{A value such that any smaller value of the objective is known to be unattainable.} for minimization problems).
It uses a \FF\ search strategy and a tree structure to navigate the search.
FDS is used in IBM CPLEX \cpsol\ \cite{ibm_cp_optimizer} (further referred to as \cpsol), and was often the algorithm responsible for proving the optimality of the solutions to the scheduling problems reported recently; see, e.g., \cite{naderia2022mixed, hauder2020resource}.
\cpsol\ is considered the state-of-the-art solver for many classes of scheduling problems (e.g, \cite{naderia2022mixed,  ABREU2022108128, ROHANINEJAD2023108958, HEINZ2022108586, LUNARDI2020105020, FATEMIANARAKI2023102770, hauder2020resource}), and although users of \cpsol\ may not notice it, FDS is an important component of this success.
We implemented FDS as described in \cite{vilim2015failure} in a new CP solver called \optsol\ with the goal of getting a deeper understanding of its notable performance and used these insights to further improve it by reducing its search tree size.

Formally, problems modeled with CP and solved by CP solvers like \optsol\ or \cpsol\ are called Constraint Satisfaction Problems (CSPs).
A CSP is characterized by a finite set of variables (the unknowns), each with its domain (a finite set of possible values) and a finite set of constraints over the variables that must be satisfied.
These constraints can take various forms, such as equations, inequalities, their boolean combinations, or special global constraints like \texttt{NoOverlap}.
We refer the reader to \cite{BRAILSFORD1999557} for a more detailed understanding of the CSP formulation.
Then, the CP solver's task is to find variable assignments from their domains (a solution) that also satisfies all given constraints.
Additionally, if an objective function is provided, the CSP becomes a Constrained Optimization Problem (COP).
In this case, the CP solver not only searches for a feasible solution but also aims to minimize or maximize the objective function.
We will further denote any CSP or COP simply as a Constrained Combinatorial Problem (CCP) because, for our purposes, the difference does not matter.

In general, CCPs are $\mathcal{NP}$-hard problems.
So, to find the optimal solution of a CCP, the CP solver needs to explore the search space exhaustively, which is where FDS comes into the picture.
To improve the speed of the FDS algorithm, we propose a two-pronged approach combining Reinforcement Learning (RL) and parameter tuning, that aims at reducing the number of created search nodes (reducing the search tree size) while still covering the whole search space. 

The remainder of this paper is structured as follows. 
In \Cref{sec:related_work}, we review the existing literature on CCP-related search algorithms, the use of machine learning for search guidance, and the combination of the two. 
In \Cref{sec:fds_as_mab}, we lay the theoretical foundation by analyzing the properties of FDS and showing that minimizing its search tree can be interpreted as the well-known RL Multi-Armed Bandit (MAB) problem \cite{bams/1183517370}.
Based on this fact, in \Cref{sec:solution_approach} we present four existing RL MAB algorithms that address the exploration-exploitation dilemma, a key aspect not considered in the original FDS.
Search-guiding (choice-selection) strategies based on those algorithms are discussed, and a problem-specific improvement called \textit{choice rollback} is proposed to increase their efficiency.
Furthermore, in \Cref{sec:param_tuning}, we discuss the tuning of the FDS parameters to improve its performance further.
Although parameter tuning is often neglected, papers like \cite{Kazikova_Pluhacek_Senkerik_2020, 10.1007/978-3-540-70881-0_24} show that it is often a significant aspect of algorithm performance.
In \Cref{sec:experimental_setup}, we describe the experimental setup and briefly introduce the \optsol\ solver whose implementation of FDS we use in the experiments.
In \Cref{sec:results_and_discussion}, we provide detailed results measuring the performance of the proposed choice-selection strategies and tuning and compare it with the state-of-the-art. 
Finally, in \Cref{sec:conclusion} we summarize the main contributions, findings, and potential directions for future research.

The main contributions of this paper are as follows.  
We propose new effective search strategies for the FDS algorithms, a state-of-the-art exhaustive search method in Constraint Programming.
We conduct a systematic exploration of the FDS parameters and provide insights into how they influence search efficiency. 
We validate our approach through extensive experiments on standard benchmark datasets for two fundamental scheduling problems, JSSP and RCPSP, demonstrating that: 
(i) the runtime of the original FDS is reduced by half, 
(ii) the enhanced FDS is 2 to 4 times faster than the state-of-the-art implementation in \cpsol, and
(iii) the state-of-the-art results in the standard benchmark sets are significantly improved.

\section{Related Work}
\label{sec:related_work}
The related literature can be grouped into three main categories based on their relevance to our research focus: (i) works that address Constraint Combinatorial Problems (CCPs) using search heuristics and algorithms without involving machine learning (see \Cref{subsec:CCPnoML}), (ii) works that use machine learning to guide search processes, such as branching decisions (see \Cref{subsec:SearchML}), and (iii) works that apply machine learning algorithms, specifically Multi-armed bandit ones, to the CCP search (see \Cref{subsec:CCPML}).

\subsection{CCP Search Without ML}
\label{subsec:CCPnoML}
There exist a multitude of search algorithms for CCPs.
They can be broadly classified into two categories: heuristic search algorithms and complete search algorithms.
Since heuristic search algorithms are not relevant for this paper, we will omit them from the review.
Regarding complete search algorithms for CCPs, the most prominent are Impact-based Search (IBS) \cite{refalo2004impact}, Activity-based Search (ABS) \cite{michel2012activity}, and FDS \cite{vilim2015failure}.
All three algorithms build a search tree and use the idea of \emph{variable ordering}; ordering variables in the branching process by their importance for the problem solution.

Among these, IBS and ABS share many conceptual similarities.
Both algorithms try to efficiently navigate towards feasible solutions, using the degree of constraint propagation at each step as a measure of the variable's future importance.
The pivotal distinction between the two is the concrete methodology employed to evaluate the importance.
The \emph{impact} in IBS represents the degree to which the assignment of a particular value to a variable reduces the domains of other variables due to the propagation process.
In contrast, \emph{activity} in ABS quantifies how many times a certain variable had its domain reduced by a propagation caused by the assignment of a value to another variable.

The FDS represents a different approach altogether.
Its main aim is to reduce the size of the search tree by efficiently navigating the search towards infeasibilities rather than solutions.
The infeasibilities found are then used to cut off parts of the solution space, making it smaller.
A similar philosophy of trying to fail quickly by focusing on (and isolating) the variables that yield infeasibilities as soon as possible is considered, for example, by \cite{LECOUTRE20091592}.
However, the primary difference of FDS compared to other existing approaches is the use of infeasibilities for a systematic search of the entire search space.

Beyond the guiding mechanisms in the aforementioned algorithms, variable ordering heuristics are specialized strategies that direct the search by determining the order in which variables are selected.
They can be categorized into two groups: Static Variable Ordering (SVO) and Dynamic Variable Ordering (DVO) heuristics.
We are going to mention only DVOs as they are usually more efficient.
One of them is a \emph{dom/wdeg} heuristic \cite{boussemart2004boosting}, which is a popular conflict-directed heuristic that combines domain propagation/reduction with a weighting of the constraints based on how well they reduce the variable domain to the empty set (domain wipeout).
Its extension \emph{wged} \cite{8995307} improves this idea by taking into account the current arity of the failed constraint, as well as the size of the current domains of the variables involved in said constraint.
Although most heuristics essentially use the arithmetic average for their ranking system, the CHS \cite{habet:hal-02090610} heuristic uses an exponential recency weighted average to estimate the change in the constraint "hardness".
This principle of accommodating the changing context of the search is something that FDS also reflects.

\subsection{General Search With ML}
\label{subsec:SearchML}
In recent years, machine learning has become frequently used in various search algorithms in different ways, but mainly (i) to guide the search with more informed decisions and (ii) to make search steps faster by estimating certain metrics that are hard to calculate exactly.
The authors of \cite{Doolaard2022} propose a way to guide the search with more informed decisions using an online machine learning approach to improve variable ordering during the search process. 
They use random forest regression to predict the quality of variable choices based on features gathered through randomized probing, enabling deeper search guidance without the overhead of explicit lookahead.
In \cite{Xu_Wu_Li_Yin_2025}, the authors propose an offline learning approach to train a selector that chooses the most suitable variable ordering heuristic for a given instance. Before the search begins, the instance is probed using multiple variable ordering heuristics, based on which the selector predicts which heuristic to apply.
In \cite{Zarpellon_Jo_Lodi_Bengio_2021}, the authors describe how a learning framework that uses parametrization of the search state is used to modulate branching decisions, leading to smaller search trees.

The machine learning estimation speeding up the search process is used by the authors of \cite{Khalil_Le_Bodic_Song_Nemhauser_Dilkina_2016}, proposing a ML surrogate function that mimics strong branching.
Strong branching is a helpful but costly process of initial search tree probing that should lead to better branching decisions close to the root of the tree.
The proposed ML surrogate function is far less computationally expensive, accelerating the algorithm.
A similar approach is proposed by \cite{BOUSKA2023990}, where ML is used to estimate the solutions of partial subproblems, eliminating the burden of complex computation by estimation.
The authors of \cite{10.1007/978-3-319-18008-3_6} use ML to replace complex or unknown components of constraint programming models with decision trees and random forests.
These learned models are embedded as constraints, allowing the solver to perform propagation over approximated knowledge.

A particular branch of ML that is closely aligned with our research is reinforcement learning (RL).
In particular, RL has proven effective for learning estimates in the context of branching heuristics, for example, in \cite{10.1007/978-3-030-78230-6_25, cappart2020combining}, where the authors leverage RL to guide branching decisions in the CP solver.
Specifically in \cite{10.1007/978-3-030-78230-6_25}, the authors use graph neural networks and a Deep Q-Network (DQN) agent.
Although their overall results are quite good considering the number of branches explored, it is stated that machine learning takes quite a lot of time.
This possibly makes simpler RL algorithms (like MAB algorithms) better candidates due to their simpler nature.
Also, they focus on value selection, while in this paper we consider both which value to assign and in which order to process the variables.
Similar approaches are considered in \cite{cappart2020combining}, where the authors note that the complexity of the approach is likely a time bottleneck of the algorithm, especially when handling larger instances, and note that reducing the complexity of machine learning might be essential.
In \cite{10.1007/978-3-319-40970-2_9}, authors consider specifically the MAB algorithm to guide the branching heuristic, but do so in a SAT solver.
For a broader perspective on various ML applications in constraint solving and search, readers can also refer to the survey paper \cite{Popescu2022}, which provides a comprehensive overview of other techniques and their respective impacts.

\subsection{CCP Search With MAB Algorithms}
\label{subsec:CCPML}
The papers discussed in this section are most relevant to our research as they apply MAB algorithms to the CCP search.
However, their use is predominantly at an external level, primarily to optimize high-level decisions such as heuristic selection, rather than being integrated directly into the search process itself.
A prominent example of this external application is the selection of the most effective search guiding heuristics from a given set. 

The approach of heuristic selection is considered in works such as \cite{Xia_Yap_2018, koriche2022best, Kletzander_Musliu_2023, Wattez2020Learning}.
In \cite{Xia_Yap_2018}, the authors focus on multiple different RL adaptive algorithms to guide the selection of heuristics, while in \cite{koriche2022best}, they use Adaptive Successive Halving and Adaptive Single Tournament to select the heuristic strategy to be used next.
In \cite{Kletzander_Musliu_2023}, authors use a novel hyper-heuristic called LAST-RL (Large-State Reinforcement Learning), and they also introduce a probability distribution for the exploration case in their epsilon-greedy policy, based on the idea of Iterated Local Search, increasing their chance to sample good chains of low-level heuristics.
In \cite{Wattez2020Learning}, the authors propose a MAB-based framework, RestartsMAB, which selects a variable ordering heuristic for each solver run and uses restart-based feedback to guide learning.
By evaluating each heuristic through pruned search tree metrics, the method outperforms both fixed heuristics and earlier bandit-based approaches.
A similar application of MAB algorithms is proposed in \cite{balafrej2015mab}, where the authors use MAB algorithms to select consistency propagation algorithms online, during the process of solving the problem instance.
The overall results show that this adaptive approach performs better than statically selecting one propagation method at the beginning of the search.

The only search heuristic that we found using the MAB algorithm internally to specifically guide the search is proposed in \cite{loth2013bandit}.
Although partially similar to our application of MAB algorithms in this paper, there are still significant differences.
The tree is constructed with Monte Carlo Tree Search and the RL algorithms (and their extensions) used are different from ours, except UCB-1, which performed poorly for our use case.
The reward calculation and other details are also quite different, and their approach is not specifically suited to the FDS algorithm.
Their results are also demonstrated on a much more limited number of instances, and no state-of-the-art results comparison is provided either.

In conclusion, the research shows that while there are papers addressing similar topics, to the best of our knowledge, no paper addresses our considered problem.
The specific applications of MAB algorithms inside the search algorithms for CCP are scarse and differ noticeably from our research.

\section{Theoretical Foundation}
\label{sec:fds_as_mab}
In this section, we will demonstrate that the selection of branches and the minimization of the search tree in FDS have the same properties as the selection of arms and the maximization of the reward sum in the MAB problem, justifying the use of MAB algorithms to minimize the FDS search tree.
Readers unfamiliar with the fundamentals of the FDS algorithm are encouraged to consult \ref{app:fds} before proceeding.

\subsection{MAB Problem and \texorpdfstring{$\epsilon$}{Epsilon}-Greedy Algorithm}
\label{sec:mab_greedy}
An RL problem is defined by actions, rewards, an agent, and the state of the environment.
When an agent chooses (plays) an action (in a state), a reward is given (possibly negative) and the state can change (in the MAB problem, the state never changes).
The goal is to maximize the sum of the rewards collected by the agent during the playing episode.

The definition of the MAB problem is the following: we have $n$ different slot machines (in casinos, these are called one-armed bandits), and our task is to select and play them with minimal loss, i.e., maximize the sum of the rewards.
Each arm draws its reward from a distribution with a different mean and a different variance (i.e., rewards are stochastic).
Therefore, we never know the exact mean value of any arm's reward, but our estimate of the arm's reward is enhanced with the number of times the arm was played.
This estimate is called the Q-value.
The main question is when to exploit already-acquired knowledge and pull the most promising arm and when to explore other arms to refine their Q-values.
In RL, this is called \emph{exploration-exploitation dilemma}.

One way to address the exploration-exploitation dilemma is the \eg\ algorithm. 
The algorithm interleaves two strategies to pick an action at each turn.
With probability $\epsilon$, the action is chosen at random (exploration).
Otherwise, the action with the highest Q-value is chosen (exploitation).
After the action is played, its Q-value is updated based on the obtained reward.

\subsection{FDS Branching as MAB Problem}
\label{subsec:fds_as_mab}
To demonstrate that branching in FDS has very similar properties to MAB problem, we will introduce a very slightly modified variant of FDS called "simpler FDS" (sFDS).
The only difference between sFDS and FDS is in the computation of $\lr$.
The reduction factor $R$ used in the original FDS $\lr$ formula (see \Cref{def:localRating} for more details) is omitted, making the new $\lr$ equation look like \Cref{def:localRatingS}.
Since $\rb$ (current \rg\ of the left/right branch) and $\ra$ (current \rg\ of choice) depend on $\lr$, we have to redefine them as well, even though their calculations do not change.
To differentiate between internal FDS and sFDS variables, we will use subscript $s$:
\begin{gather}
   \lrs :=
   \begin{cases}
      0 &\text{\rmfamily if branch is infeasible}, \\
      1 &\text{\rmfamily otherwise},
   \end{cases}
  \label{def:localRatingS} \\
  \rbs := (1-\alpha) \cdot \rbs + \alpha \cdot \lrs,
  \label{def:ratingupdS} \\
  \ras := \rpls + \rmis.
  \label{def:ratingS}
\end{gather}

Since we will be examining the behavior of a single search-tree expansion, restarts and no-good recording normally used in FDS are not considered.
We also exclude a strong branching mechanism as the goal is to isolate and examine the properties of FDS branching specifically.
Now note that choice selection in sFDS is analogous to arm selection in the MAB problem: at each search node where infeasibility was not detected, we have to pick one choice to branch upon.
In MAB, we prefer actions with high rewards, while in sFDS, we prefer choices with small rating values.
So, we define localReward (reward for one branch) as a negation of obtained $\lrs$ :
\begin{gather}
    \text{localReward} = - \lrs,
    \label{def:rewardPenalty}
\end{gather}
making the reward for the choice:
\begin{gather}
   \text{reward} :=
   \begin{cases}
      0 &\text{\rmfamily if both branches are infeasible}, \\
      -1 &\text{\rmfamily if one branch is infeasible}, \\
      -2 &\text{\rmfamily if both branches are feasible}.
   \end{cases}
\end{gather}
This makes \Cref{def:ratingupdS} equivalent to the standard Q-learning equation:
\begin{gather}
    Q(c) := Q(c) + \alpha \cdot (\text{reward} - Q(c)),
    \label{def:Qvalue}
\end{gather}
where $Q(c)$ represents the Q-value of choice which is inverse to its $\ras$.
Thus, by transforming $\lrs$ to reward (and therefore $\ras$ to $Q(c)$), it becomes apparent that both are very similar.

If sFDS has the same properties of $\ras$/$Q(c)$ calculation as the MAB problem, the remaining question is what is being optimized by maximizing the sum of rewards, i.e., minimizing the sum of $\lrs$s?
As the penalty sum in sFDS increases with every newly opened search node that does not fail, it directly corresponds to the search tree's internal node count.
Since half of the nodes in a binary search tree of sFDS are internal, the penalty sum is proportional to the search tree size.
Thus, minimizing the penalty sum is the same as minimizing the search tree size, which is exactly what we want the sFDS (and FDS) algorithm to do.

The behavior of the sFDS algorithm is actually the same as that of the MAB \eg\ algorithm (with $\epsilon=0$) which implicitly minimizes the search tree.
Therefore, other more complex state-of-the-art MAB algorithms are suitable candidates to extend the FDS functionality and improve its performance.

However, we need to address two important aspects of sFDS that were not mentioned before and make sFDS different from the classical stationary (i.e. fixed number of arms) MAB problem:
\begin{enumerate}
    \item 
    \label{point:one}
    The set of choices available in sFDS is different based on the current search state (position in the search tree) as some choices might have already been used.
    This would be analogous to changing the available set of arms in the MAB problem.
    Although this is not possible in the classical MAB problem, other variants exist.
    In particular, there is \emph{arm-acquiring} MAB problem, where the set of arms grows.
    While this variant has scarcely been studied, \cite{Mahajan2008} shows that in terms of the optimal strategy, it behaves similarly to the general MAB problem.
    \item
    \label{point:two}
    Q-values of arms in the MAB problem are independent, and their reward distribution is unchanging.
    This is not true for sFDS, as the outcome of a choice is quite dependent on the preceding choices in the search tree.
    Therefore, rewards can change dramatically as the algorithm progresses.
    In RL, such an action behavior is called \emph{non-stationary}, and such MAB problem variants are also studied.
    To demonstrate that the rating of the choices exhibits the \emph{non-stationary} behavior, we provide an experiment in \ref{app:fixrat}.
    However, in \cite{sutton2018reinforcement}, it is shown that the classical MAB algorithms are also efficient for \emph{non-stationary} MAB.
\end{enumerate}
Similar observations as \Cref{point:one,point:two} were also made in \cite{10.1007/978-3-319-40970-2_9}, where the authors used the MAB framework in the SAT solver.
Their success further demonstrates that these aspects do not have to be limitations to the application of MAB algorithms.
However, it is important to point out that the guarantees (regret bounds, optimal action, etc.) given by algorithms in the classical MAB problem scenario do not have to hold in our case.

Since we showed the correspondence of sFDS with the MAB problem and \eg\ algorithm, we can explore how the parameters of \eg\ algorithm relate to FDS and what knowledge about them can be transferred.
In particular, we will focus on the notion of an exploration-exploitation dilemma that is regulated through parameters $\boldsymbol{\alpha}$, an initial rating of every branch (called \textbf{$\dr$}), and especially by $\boldsymbol{\epsilon}$.

Let us start with $\boldsymbol{\alpha}$.
While the value of $\alpha$ used in RL applications varies depending on the task, it usually does not exceed $0.1$, which means that the last reward obtained is \SI{10}{\percent} of the newly calculated Q-value.
However, for non-stationary problems, even higher $\alpha$ is often necessary to increase the importance of the recent observations and adapt to the new reward distributions, ensuring the exploration of newly emerged promising actions.
This is why in FDS, a "hybrid $\alpha$" is used; see \ref{subsec:factor_alpha} for more details.
With $L=30$, the hybrid $\alpha$'s value is between $0.5$ and $0.03$, where it stabilizes after $L$ updates of that particular rating.

Next, consider \textbf{$\dr$}.
It is known that the \eg\ algorithm is particularly sensitive to initial Q-values, which corresponds to our experience with FDS and branch $\dr$.
In \cite{sutton2018reinforcement}, a strong argument is made for the \emph{optimistic initial values}, meaning the initial Q-values are set to a value higher (lower in the case of our rating) than the expected average reward.
This ensures exploration of all available actions (choices) at the start, unless very good ones are found right at the beginning.
However, in the original FDS, \emph{realistic initial values} were used instead, which makes choices with values above average likely to be exploited right away.
Thus, decreasing $\dr$ from the original $1.0$ to $0.3$, making the unexplored choices more likely to be selected, noticeably improved the performance of FDS.

Lastly, there is $\boldsymbol{\epsilon}$, which represents the chance to pick an exploratory choice instead of the best one.
As already stated, the original FDS did not explore ($\epsilon=0$).
However, the existing literature and experiments on the exploration-exploitation dilemma make it clear that exploration is necessary.
The best amount of exploration is problem-dependent, but papers like \cite{DBLP:journals/corr/KuleshovP14} provide an overview of how degrees of exploration affect reward maximization in the MAB problem.
Furthermore, for non-stationary problems, increasing $\epsilon$ is important as finding newly promising actions can depend on the exploration as well.
This is why, for FDS, the parameter tuning in \Cref{sec:param_tuning}, suggested a relatively high value of $\epsilon=0.10$ as the best option.

\subsection{Effects of Rating Learning}
\label{subsec:effects_learning}
To demonstrate the effect of learning and the importance of addressing the exploration/exploitation dilemma, we show how $\epsilon$-greedy with \SI{3}{\percent} exploration learns the ratings and gradually reduces the search tree size.
To evaluate that, the ratings learned in one completed run are passed as initial ratings to the following one, measuring if runtime was shortened.
Search restarts were not used.

For this experiment, we used 55 JSSP and 80 RCPSP instances, each paired with an infeasible upper bound.
From the instances considered in \Cref{sec:experimental_setup}, we selected those with infeasibility proofs of sizes between the $10,000$ and $250,000$ branches so that the proofs are neither trivial nor too expensive.
Still, the instances are not trivial, as unguided random choice selection would fail to obtain an infeasibility proof in $50,000,000$ branches for any of the tested instances.

\begin{figure}[ht]
    \centering
    \begin{minipage}{0.48\textwidth}
        \resizebox{\textwidth}{!}{
            \includegraphics[width=\linewidth]{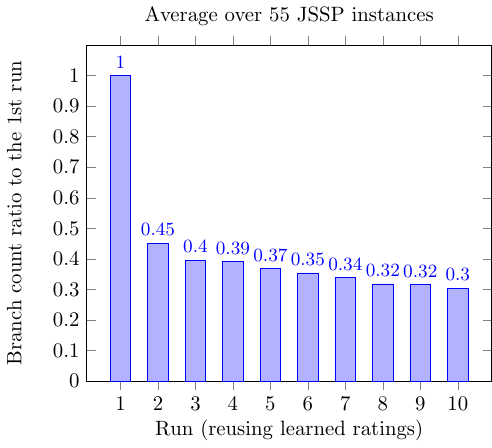}
        }
    \end{minipage}
    \hfill
    \begin{minipage}{0.48\textwidth}
    \resizebox{\textwidth}{!}{
        \includegraphics[width=\linewidth]{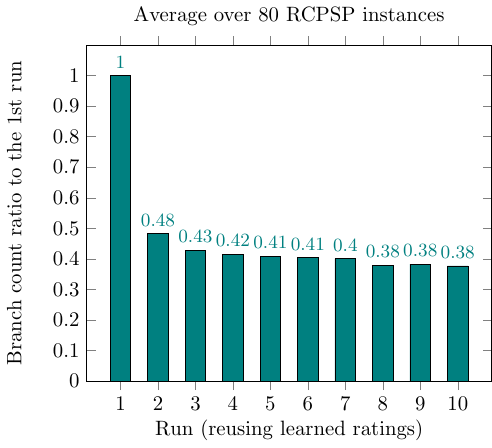}
        }
    \end{minipage}
    
    \caption{Bar charts showing gradual reduction of explored branches needed to prove instance infeasibility, when choice ratings from one run are passed as initial ratings to the next one on the same instance. 
    }
    \label{fig:barchart}
\end{figure}

\begin{figure}[ht]
    \centering
    \begin{minipage}{0.48\textwidth}
        \includegraphics[width=\linewidth]{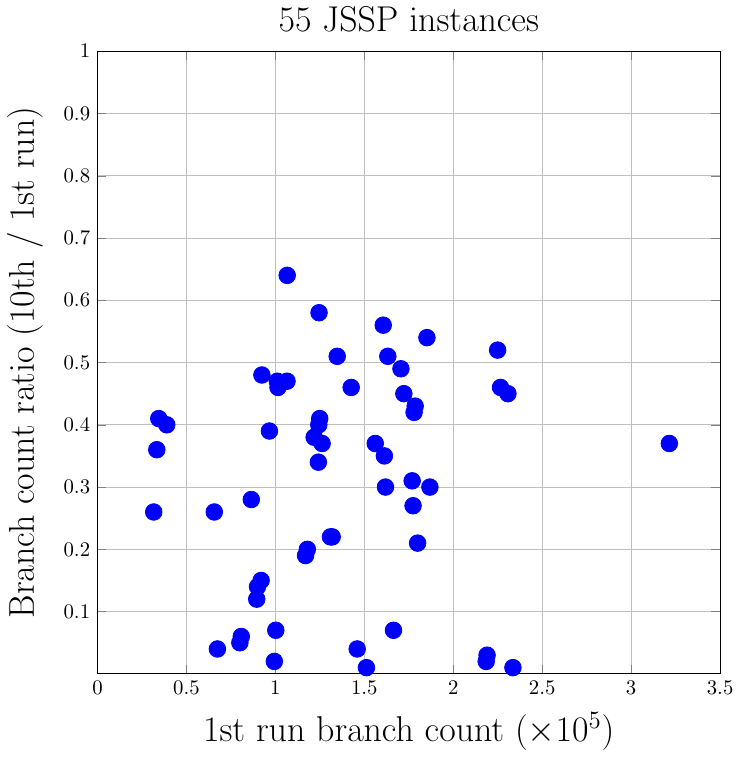}
    \end{minipage}
    \hfill
    \begin{minipage}{0.48\textwidth}
        \includegraphics[width=\linewidth]{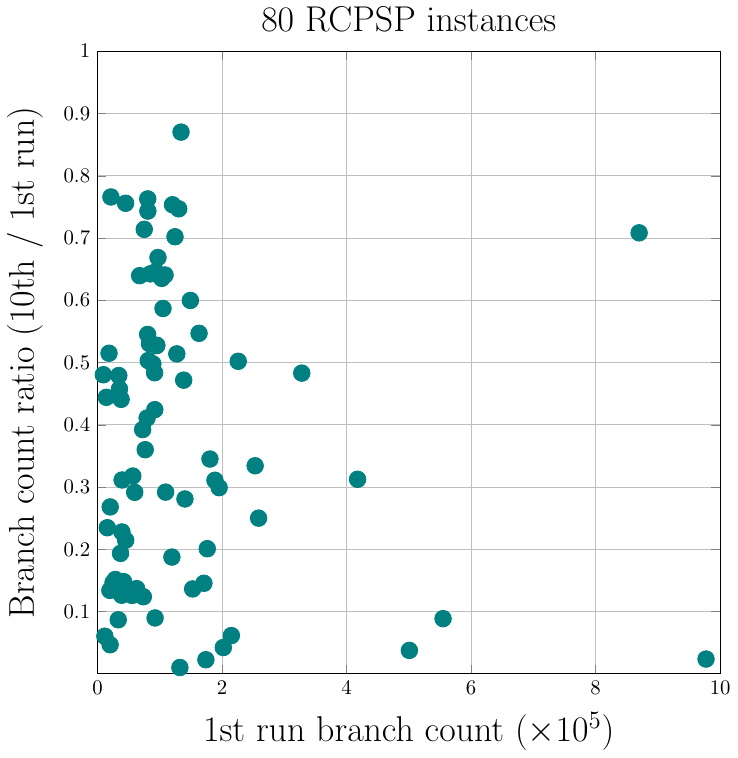}
    \end{minipage}

    \caption{Scatter plots showing the ratio between explored branches in the 1st run and the 10th run of the algorithm with choice ratings being passed between all runs.
    }
    \label{fig:scatterplot}
\end{figure}

FDS was run ten times in a row for all instances, each time reusing the ratings from the previous run.
In the first run, the ratings are initialized by the parameter \dr\ as usual.
\Cref{fig:barchart} shows the average ratio of branches explored in increasingly repeated runs compared to the first run of the algorithm with the individual instance measurements provided in \Cref{fig:scatterplot}.
It can be seen that, on average, the second run takes less than \repeatedSavings\ of the branches, both for JSSP and RCPSP.
The number of branches further drops to \repeatedLastSavingsJSSP\ (JSSP) and \repeatedLastSavingsRCPSP\ (RCPSP) in the tenth run repetition.
The conclusion is clear: learning the ratings undoubtedly leads to minimizing the search tree size (even though in practical application we only run the algorithm once to find the solution).

\section{Solution Approach}
\label{sec:solution_approach}
This section provides a detailed discussion of the MAB algorithms employed, referred to as choice-selection strategies, and the associated parameter tuning process.

\subsection{Choice-Selection Strategies}
\label{sec:choice_selection}
To address the exploration-exploitation dilemma in FDS, we chose the following 4 prominent MAB algorithms to use and extend: \emph{\eg}, \emph{Boltzmann exploration} (also known as \emph{Softmax}), \emph{UCB-1} and \emph{Thompson sampling}.
Their specific applications to FDS will be called \emph{choice-selection strategies}.
All of these strategies compute their Q-values in the same way as we have described in \Cref{def:Qvalue}, but the way they select the action is different.
A quick overview of the strategies follows (not including previously explained \eg).
More details can be found, e.g., in \cite{DBLP:journals/corr/KuleshovP14} and \cite{NIPS2011_e53a0a29}.
Note that the strategies discussed below extend the choice selection logic of FDS (\SC\ function on \Cref{code:beginDecided} in \Cref{alg:fds} in the provided pseudocode).

\paragraph*{Boltzmann Exploration}
In Boltzmann's exploration, every action is assigned a probability of being picked.
Probabilities are computed from Q-values using a \emph{softmax} function.
The action is then selected randomly based on these computed probabilities.
The probability of selecting action $a$ is:
\begin{gather}
  P(a) = \frac{ e^{Q(a) / \tau} } { \sum_{a' \in A}{ e^{Q(a') / \tau } } },
\end{gather}
where the sum is over the set of all available actions $A$ and $\tau > 0$ is a \emph{temperature} parameter.
When $\tau$ is high, all actions have almost the same probability of being selected.
As $\tau$ goes to zero, the best action is almost always selected.
In our tests, we used the fixed $\tau=1$.
In general, better actions always have a higher probability of being selected, but probability-based
selection ensures a balance between exploration and exploitation.

\paragraph*{UCB-1}
UCB-1 always picks the best action $a_t$ but not according to the Q-values, but based on the following equation:
\begin{gather}
    a_t = \text{argmax}_{a \in A} \left(Q(a) + \sqrt{ \frac{2 \ln t}{N(a)} }\right),
\end{gather}
where $t$ is the number of all actions taken and $N(a)$ the number of times a specific action $a$ was taken.
Thus, both action's quality and its exploration are considered.
As a result, actions that were taken only a few times will be picked sooner or later, again balancing exploration with exploitation.

\paragraph*{Thompson Sampling}
One issue that none of the strategies mentioned above addresses is the uncertainty of the action's expected value estimate.
Thompson sampling addresses this by modeling the action's expected value as a distribution instead of one concrete number.
Then, when the next action to play is selected, a value is drawn from every action's distribution, and the action with the best-sampled value is selected.
In our case, we used the Gaussian distribution within the Thompson sampling algorithm.
More details about the algorithm can be found in \cite{NIPS2011_e53a0a29}.

\subsubsection{Strategy Hybridization}
As shown in \cite{DBLP:journals/corr/KuleshovP14}, simpler strategies often provide better results for the MAB problem.
Especially in situations where greedy action is very good or when we have already identified the best action, more complicated strategies become less effective (often due to their persistent effort to explore). 
In FDS, this occurs when a \emph{closing choice} (a choice having infeasibilities in both branches) is found.
Usually, \emph{closing choice} will remain "closing" even after a backtrack, so exploiting it repeatedly makes sense.
Because \emph{closing choice} usually has the best rating, the original greedy strategy in FDS is quite effective in case such a choice is found.

This observation led to the idea of combining the Boltzmann exploration, UCB-1, and Thompson sampling strategies with greedy choice selection in the same way as the \eg\ strategy combines the selection of a random choice with the greedy one.
This approach ensures that, in cases where a choice that is clearly best exists, we exploit it.
The resulting hybrid strategies select the exploration choice according to one of the three explained strategies (since exploration is embedded in them).
The greedy choice (the choice with the best rating) is otherwise selected.
These strategies are called \bg\ (for Boltzmann exploration), \ug\ (for UCB-1) and \tg\ (for Thompson sampling), denoting the strategy that provides the exploration choice (similarly to \eg\ that provides the random exploration choice).

Thus, we can think of it as having 4 non-hybrid strategies: original pure greedy (\eg\ with $\epsilon=0$), Boltzmann exploration, UCB-1, and Thompson sampling, and 4 hybrid (alternating) strategies: \eg, \bg, \ug, and \tg.

\subsubsection{Choice Rollback}
\label{subsec:improving_the_strategies}
While exploration is the key to an efficient search for the best choices, it can be quite expensive in the context of search tree minimization.
Imagine a situation where a random choice is selected instead of the best one.
This random choice can create two new branches without contributing anything to make the two new subproblems easier to solve, essentially doubling the size of the search tree.
This makes the exploration cost very high in the context of FDS.

A hint of exploration incorporated in the original FDS is a strong branching mechanism, but it is limited in its scope; see \ref{subsec:strong_branching} for more details.
However, its important property is that it does not increase the size of the tree.
In fact, it does the opposite by carefully exploring which choice might be the best.
Based on that idea, we propose a new way of exploration: first, evaluate a choice in the current context and update its rating without adding it to the search tree, then decide if said choice should be used for real branching.
It clearly makes sense to use the choice for branching if both branches fail but also if only one branch fails, as the search tree size does not increase, and useful domain propagation might occur.
If this is not the case, we do not use the choice for branching, but its rating is still updated, so the effect of exploration is not completely lost.
We call this process \emph{choice rollback}.

Note that \emph{choice rollback} is applied only to exploratory choices.
This means that the algorithm has to distinguish between an exploratory and an exploitatory action.
This is possible in the case of hybrid (alternating) strategies that we consider, but for pure Boltzmann exploration, UCB-1, and Thompson sampling strategies, it is hard to determine if the selected action is (mostly) exploratory or (mostly) exploitatory.
For this reason, we did not consider choice rollback for pure strategies.

It is important to note that even choice rollback introduces some overhead.
Its cost is essentially bounded by the number of exploratory choices, since all may be rolled back in the worst case.
If \SI{10}{\percent} of the choices are exploratory, runtime could increase by up to \SI{10}{\percent}.
Still, even rolled-back choices update their ratings, obtaining information about their quality which is helpful.
Moreover, measured across 20 randomly selected JSSP and RCPSP instances, only about \SI{70}{\percent} of the exploratory choices are rolled back at a \SI{10}{\percent} exploration rate, meaning \SI{30}{\percent} of them still actively contribute to the search tree.
While no formal guarantee of runtime improvement can be given, the results in \Cref{tab:mab_comparison} show that incorporating choice rollback consistently yields better performance across all tested choice-selection strategies and exploration rates.
The benefit of using choice rollback is also evident on the best choice-selection strategy by comparing row 3 with row 4 in \Cref{tab:all_comparison}.

\subsection{Parameter Tuning}
\label{sec:param_tuning}
In the previous section, we proposed extensions to the FDS algorithm, improving its efficiency.
However, these extensions come with parameters that can be adjusted to further affect performance, for example, the value of $\epsilon$, the value of \dr, etc.
Not only that, the FDS (and its particular implementation in \optsol\ solver that we used) already has many parameters that can be considered as well.

In the parameter-tuning process, we considered a total of 24 FDS-related parameters (out of 27 available at the time) that exhibited a measurable impact on the solution process when their values were changed: AdditionalStepRatio, BothFailRewardFactor, InitialRating, Epsilon, EpsilonDecay, EventTimeInfluence, FixedAlpha, InitialRestartLimit, LengthStepRatio, MaxCounterAfterRestart, MaxCounterAfterSolution, MaxInitialChoicesPerVariable, PenaltyNotFailed, PresenceStatus-\hspace{0pt}Choices, RatingAverageComparison, RatingAverageLength, ResetRestartsAfterSolution, RestartGrowthFactor, RestartStrategy, StartVariableAlphaN, StrongBranchingCriterion, StrongBranchingDepth, StrongBranchingSize, UseNogoods.
Their explanation is provided in \ref{app:fds_params}.

The primary objective was to identify suitable configurations of parameter values for the JSSP and RCPSP problem.
While such configurations might generalize to other scheduling problems with similar properties, this cannot be asserted with certainty and falls outside the scope of this paper.
However, the identified promising parameter values for JSSP and RCPSP may serve as a valuable starting point for further tuning on other combinatorial problems.

The main challenge was to efficiently identify impactful parameters, explore their mutual effects, and determine their optimal values within a tractable timeframe, leading to the following multistep tuning process:
\begin{enumerate}
    \item \textbf{Preprocessing:}
    To roughly estimate the best values for the parameters, the bounded continuous domains were uniformly sampled, effectively emulating a grid search.
    For discrete bounded domains (such as boolean), all possible values were tested.
    For unbounded domains, sampling was restricted to practical ranges.
    \item \textbf{Exploration:}
    We exhaustively evaluated all single, double, and limited triplet combinations of the 24 selected parameter values using easy-to-solve instances.
    These instances were created by an additional upper bound constraint that made them solvable in under 10 seconds. 
    The goal was to estimate how each parameter affects the search on its own but also how it influences the others.
    \item \textbf{Selection:}
    Only 200 best configurations from the previous step were kept, while same parameter configurations with dominated values of re-ocurring parameters were eliminated.
    Next, the evaluation was performed using harder instances (solvable in \SI{30}{\second} time limit).
    As a result, 11 key parameters occuring in multiple configurations were identified for further exploration, while the remaining ones were disregarded due to their negligible overall impact.
    \item \textbf{Tuning:}
    We evaluated the best parameter-value combinations from the previous step, containing the 11 parameters using the hardest instances (solvable in \SI{150}{\second} time limit).
    Additionally, we also used Optuna Hyperparameter Optimization Framework on these parameters, an automatic parameter tuner \cite{optuna_2019}, which automatically samples parameter domains, identifies promising regions, and estimates parameter importance.
\end{enumerate}

The results revealed 4 parameters with the clear best settings:
\begin{enumerate}
    \item LengthStepRatio (LSR) --  a float between 0 and 1, representing the relative length of interval split shift between two choices (the distance between two pivots) on the same variable, compared to the whole interval variable length,
    \item UniformChoiceStep (UCS) -- a boolean representing whether LSR is calculated uniformly (based on the average of all intervals) or non-uniformly (based on the length of each individual interval variable),
    \item RatingAverageComparison (RAC) -- a boolean representing whether $\av$ is used,
    \item RatingAverageLength (RAL) -- a positive integer representing the $L$ in the $\alpha$ calculation. Values between 5 and 100 were tested in tuning process.
\end{enumerate} 

The most important parameter is LSR because it is both extremely sensitive and quite dependent on the type of problem at hand, since the interval size of the branch greatly affects the ability to propagate and close subtrees.
For JSSP, its value must be between 0.6 and 0.85, and for RCPSP, between 0.3 and 0.55.
Otherwise, the performance drops by an order of magnitude or more on some of the benchmark instances.
To understand how it works together with the UniformChoiceStep parameter, we provide an example in \ref{app:gic}.
For the remaining parameters, the best setting is to use uniform intervals, omit $\av$ as explained at depth in \ref{app:avd}, and set $L=30$.

The remaining 7 key parameters were explored more thoroughly.
\Cref{tab:parameter_comparison} contains the importance and best setting for each parameter estimated by Optuna \cite{optuna_2019}.
Note that the values presented are independent of our grid search tuning.
\begin{table}[ht]
\centering
    \includegraphics[width=\linewidth]{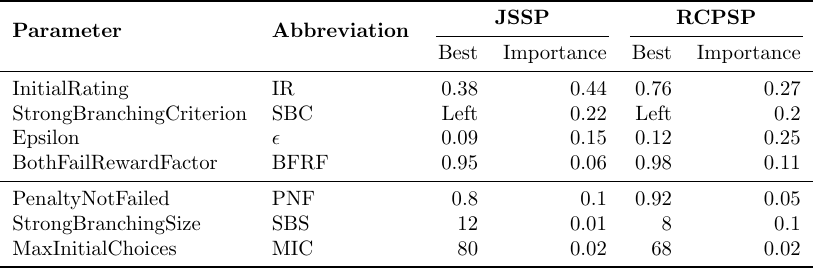}%
    \caption{The parameters with their best setting and their relative importance compared to others, given by Optuna for JSSP and RCPSP problems.
    The concrete best setting of each parameter can be dependent on other parameter values.}
    \label{tab:parameter_comparison}
\end{table}
The most important takeaways with respect to the four parameters in the upper part of \Cref{tab:parameter_comparison} (the more influential ones) are the following.
\begin{itemize}
    \item $\dr$ (values between 0 and 2 were sampled in tuning) should be set according to the idea of \emph{optimistic initial values} \cite{sutton2018reinforcement}.
    \item When deciding on which choice to branch after evaluating, strong branching should only consider $\lr$ of the \emph{left} branch, ignoring $\lr$ of the right branch.
    \item It pays off to pick a surprisingly high (roughly \SI{10}{\percent}) of the exploratory choices (with choice rollback).
    \item In case both branches of choice produce \texttt{Fail}, it is worth multiplying their ratings by a factor represented by parameter BFRF to further increase choice's chance of being picked.
\end{itemize}

The knowledge and results obtained from the tuning process were used to set the parameter values in the final configuration.
Enhanced FDS with the best-tuned parameters is \jsspTuning\ times faster on the JSSP instance set and \rcpspTuning\ times faster on the RCPSP than without tuning, as can be seen by comparing row 4 and row 5 in \Cref{tab:all_comparison}.

\section{Experimental Setup}
\label{sec:experimental_setup}
Two sets of \jsspCompleteHardSet\ JSSP and \rcpspCompleteHardSet\ RCPSP instances were created.
The JSSP instance set contains instances from \cite{TAILLARD1993278}, \cite{DEMIRKOL1998137}, \cite{10.2307/2632051}, \cite{10.2307/2632676} and \cite{inproceedings} instance sets, picking only those instances, where the original FDS was not able to prove (find) optimal makespan of the instance in \SI{150}{\second}.
Since RCPSP has a larger set of standard instances, our instance set contains the hardest instances from the PSPLIB \cite{KOLISCH1997205} J60, J90 and J120 set of instances.
The number of instances per each of the J60, J90, and J120 sets was proportional to the number of sufficiently hard instances in that set.

\begin{figure}
    \centering
    \includegraphics[width=\linewidth]{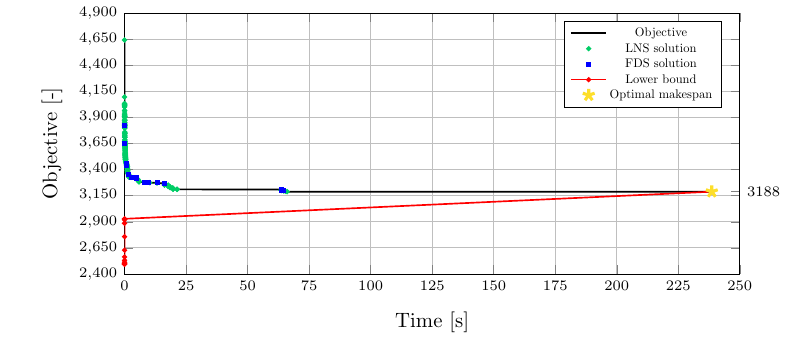}
    \caption{
        Graph demonstrating the cooperation of LNS and FDS in solving process on (formerly) open instance "Dmu07\_rcmax\_20\_20\_7" from \cite{DEMIRKOL1998137}.
        LNS and FDS were executed in parallel threads and exchanged solutions.
        Solutions found by LNS are denoted by green, and solutions found by FDS are denoted by blue color.
        The gradual improvement of the best incumbent solution (black) and lower bound (red) converge to the provably optimal solution denoted by a golden star (3188).
        The previous state-of-the-art lower bound in literature was 3051.
    }
    \label{gra:run}
\end{figure}

The primary objective of the experiments is to demonstrate that by applying reinforcement learning MAB algorithms, the performance of the FDS algorithm in \optsol\ is improved.
This also aims to validate the original hypothesis that the lack of exploration in the original FDS significantly impacts its performance.
In particular, we are interested in the typical use case of FDS, i.e., in the cooperation with the other search algorithm called Large Neighborhood Search (LNS) used in the \optsol\ solver.
To this end, we introduced an additional constraint in all instances, limiting their solution makespan to a lower value than the optimal one, making all instances infeasible.
The reasoning can be summarized by the following three points:
\begin{itemize}
  \item
  For FDS, it is the natural scenario to have an upper bound on the solution because it is provided by LNS; see \Cref{gra:run} illustrating a typical \optsol\ execution.
  Moreover, the proof of optimality/infeasibility is usually the hardest part (in \Cref{gra:run}, from the 66th second, we already have the optimal solution, so \SI{70}{\percent} of runtime is the optimality proof).
  This means that proving that a value one lower than the optimal makespan is infeasible is very often the largest part of the solving process for FDS (and also for the solver itself).
  \item 
  To reduce results variability by ensuring that no solution is randomly found during the search, influencing the runtime by having an upper bound constraint reducing the problem's feasible solution space.
  \item
  To tweak and even out the hardness of testing instances by changing the maximum allowed makespan (lowering it makes the instance easier).
\end{itemize}

Furthermore, these sets of instances were split into two.
The development sets of \jsspTrainingSet\ JSSP and \rcpspTrainingSet\ RCPSP instances were actively used during development to measure progress and tune the parameters (see \Cref{sec:param_tuning}), and the evaluation sets of \jsspTestingSet\ JSSP and \rcpspTestingSet\ RCPSP instances were used to measure all results provided in \Cref{sec:results_and_discussion}.

A secondary objective pursued in the experiments was improving current state-of-the-art lower bounds on yet open\footnote{Instance without existing proven optimal solution.} instances, proving that search speedup achieved by modified MAB algorithms is not due to the poor performance of the original FDS, but rather due to the state-of-the-art capabilities of the enhanced FDS.
Thus, in \Cref{subsec_sotacomp}, all \jsspAll\ open JSSP instances from \cite{TAILLARD1993278}, \cite{DEMIRKOL1998137}, \cite{10.2307/2632051}, \cite{10.2307/2632676} and \cite{inproceedings} and all \rcpspAll\ open RCPSP instances from \cite{KOLISCH1997205} are used instead of the development and evaluation tests described above.

All tests were performed using \optsol\ solver 0.8.5 \footnote{The solver is available for download at \href{www.scheduleopt.com}{www.scheduleopt.com}.} \cite{optalcp}.
\optsol\ is a new modern CP solver focused on scheduling problems implementing all major scheduling constraints such as precedences, unary (No-overlap) or cumulative resources, arithmetic expressions, etc. while also supporting optional interval variables, alternative constraints, or execution costs.
All measurements were performed using a single processor core to assess the performance of the FDS rather than the effectiveness of the parallelism.
We also compared \optsol's FDS with the FDS in \cpsol\ 22.1.
To ensure the same conditions, \cpsol's parameters were set to \texttt{FailureDirectedSearchEmphasis=1} and \texttt{Workers=1} (ensuring only one FDS worker is used).
All experimental results were obtained on an Intel Xeon 4110 processor running at 2.1GHz.
All sets created (development, evaluation, and open instances) for both problems and detailed results can be found in the GitLab repository \cite{fds_experimental_results}.

\section{Results and Discussion}
\label{sec:results_and_discussion}
This section contains the experimental results divided into 3 subsections.
In \Cref{subsec:choice-selection}, we present the comparison of choice-selection strategies and discuss the implications stemming from the results.
In \Cref{subsec:overall}, we show the effect of applying the best choice-selection strategy and compounding it with parameter tuning.
This best obtained configuration (row 5 in \Cref{tab:all_comparison}) is compared to the original FDS algorithm also implemented in \optsol\ (row 1 in \Cref{tab:all_comparison}) and to \cpsol's FDS (row 6 in \Cref{tab:all_comparison}).
Finally, in \Cref{subsec_sotacomp}, we present the comparison of our obtained lower bounds with the state-of-the-art ones on all \jsspAll\ open JSSP instances from \cite{TAILLARD1993278}, \cite{DEMIRKOL1998137}, \cite{10.2307/2632051}, \cite{10.2307/2632676} and \cite{inproceedings} set and on all \rcpspAll\ open RCPSP instances from \cite{KOLISCH1997205}.
We consider the results presented in \Cref{subsec:overall} and \Cref{subsec_sotacomp} to be the key achievements of this paper.

\subsection{Choice-Selection Strategies}
\label{subsec:choice-selection}
Given the choice-selection strategies described in \Cref{sec:choice_selection}, we considered 20 possible configurations in the testing process.
The 4 non-hybrid tested strategies were pure greedy, Boltzmann exploration, UCB-1, and Thompson sampling.
The 4 hybrid strategies tested were considered with and without \emph{choice rollback} and with two different levels of exploration rate, resulting in 16 configurations.
The overall knowledge obtained by this extensive testing is the following:
\begin{enumerate}
    \item The greedy selection in hybrid strategies is vital.
    \item Less complex strategies (\eg\ and \bg) perform better, with \bg\ performing the best.
    \item \emph{Choice rollback} is very efficient for all algorithms.
\end{enumerate}

Detailed results are provided in \Cref{tab:mab_comparison}.
All strategies were tested using the parameter setting from row 3 in \Cref{tab:all_comparison} (except $\epsilon$).
Since the pure (non-hybrid) strategies performed strictly worse than their hybrid counterparts, they are not presented in the comparison table.
This again demonstrates that strong exploitation is important in FDS once the best choice is located.
Please note that the number of executed branches per second can differ for each row in \Cref{tab:mab_comparison} due to the constraint propagations in search nodes, which can take different amounts of time.
The added complexity by using choice-selection strategies was measured to be negligible.
Finally, the verification of the statistical significance of the results obtained is provided in \ref{app:stat_anal}.

\begin{table}[ht]
\centering
    \includegraphics[width=\linewidth]{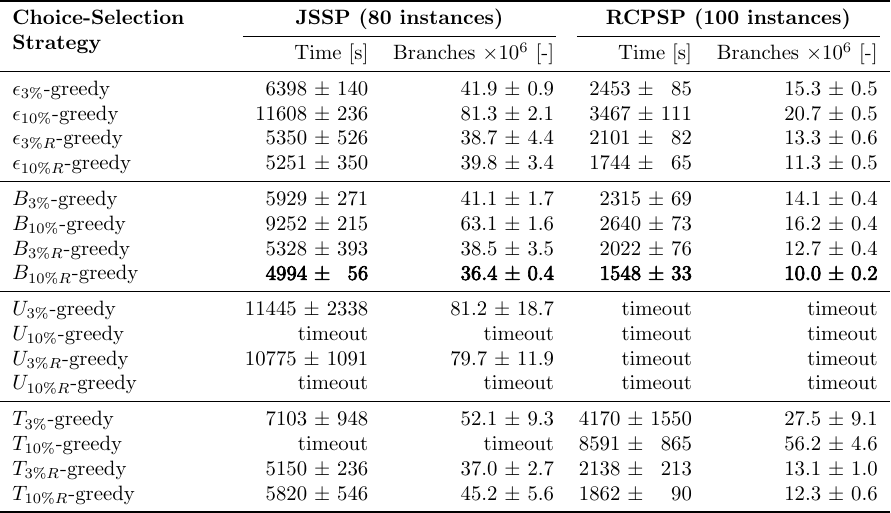}
    \caption{
        Comparison of hybrid choice-selection strategies considering two exploration rates (\SI{3}{\percent} and \SI{10}{\percent}) and the \emph{choice rollback} denoted by subscript $R$.
        The time and explored branches are 7-run averages (a different seed for each run) over sums of all tested instances.
    }
    \label{tab:mab_comparison}
\end{table}

\subsection{Overall Evaluation}
\label{subsec:overall}
\Cref{tab:all_comparison} contains the general comparison of runtime (and branches explored), showing gradual improvements achieved by the reinforcement learning and parameter tuning of \optsol's FDS.
All instances were solved, which means they were proved infeasible under a given upper bound.
The original FDS algorithm (row 1) was implemented in \optsol\ as described in \cite{vilim2015failure}.
The comparison shows that the enhanced FDS (row 5) is \jsspToOldFDS\ times faster on JSSP and \rcpspToOldFDS\ times faster on the RCPSP benchmarks than the original FDS (row 1), demonstrating an indisputable improvement.
It is also \jsspToCPOpt\ times faster on JSSP and \rcpspToCPOpt\ times faster on RCPSP than the current state-of-the-art FDS in IBM \cpsol\ 22.1 \cite{ibm_cp_optimizer} (row 6).

While the difference between \optsol's and \cpsol's FDS execution times can be affected by implementation efficiency, a comparison of the explored branches shows that our enhanced FDS is superior to \cpsol's FDS.
Finally, the verification of the statistical significance of the results obtained is provided in \ref{app:stat_anal}.

\begin{table}[ht]
\centering
    \includegraphics[width=\linewidth]{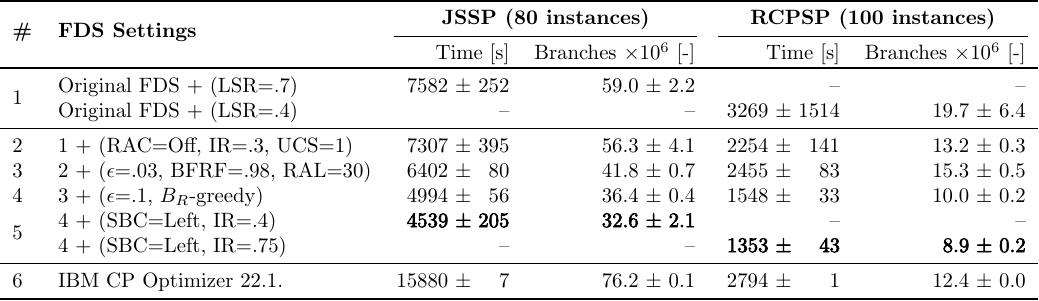}
    \caption{
        (\#1) Original FDS \cite{vilim2015failure} implemented in \optsol,
        extended with 
        (\#2) basic parameters tweaked,
        (\#3) hybrid $\alpha$ and \eg\ exploration,
        (\#4) best choice-selection strategy with choice rollback and increased exploration ($\epsilon$),
        (\#5) Optuna-based best parameter settings.
        (\#6) FDS in IBM \cpsol\ 22.1.
        A row number with a plus sign denotes inherited settings.
        The time and explored branches are 7-run averages (a different seed for each run) over sums of all tested instances.
        Abbreviations of parameters are described in \Cref{sec:param_tuning}.
    }
    \label{tab:all_comparison}
\end{table}

\subsection{State-of-the-Art Lower Bound Improvement}
\label{subsec_sotacomp}
As described in \Cref{sec:experimental_setup}, we used all \jsspAll\ open JSSP and all \rcpspAll\ open RCPSP instances from the used JSSP \cite{TAILLARD1993278}, \cite{DEMIRKOL1998137}, \cite{10.2307/2632051}, \cite{10.2307/2632676}, \cite{inproceedings} and RCPSP \cite{KOLISCH1997205} instance sets to compare our enhanced FDS with the state-of-the-art lower bounds.
For each open instance, we calculated a lower bound as a value equal to one unit above the highest makespan proven to be infeasible within \SI{900}{\second} using the enhanced FDS.
Then, we compared these lower bounds with the state-of-the-art ones from the literature on all the aforementioned instances.
For JSSP, the values were sourced from \cite{optimizizer_jobshop} and \cite{jjvh_jobshop}; for RCPSP, they were sourced from \cite{ugent_rcpsp}.
The comparison is provided in \Cref{tab:sota_comparison}.
Please note that the lower bounds reported in the literature could have been obtained using a variety of methods (see, for example, the original lower bound calculation in \cite{TAILLARD1993278}), under varying time limits (e.g., \cite{brinkkotter2001} reports runtimes extending to several hundred thousand seconds) and on significantly different hardware.
Therefore, we make no claims about the overall efficiency of our method compared to those of others.
\begin{table}[ht]
\centering
    \scalebox{0.82}{
        \begin{tabular}{lrrrrrrrrr}
            \toprule
            \multirow{2}{0em}{\textbf{Result}} & \multicolumn{6}{c}{\bf JSSP} & \multicolumn{3}{c}{\bf RCPSP} \\
            \cmidrule(lr){2-7}\cmidrule(lr){8-10}
            \multicolumn{1}{c}{} & ABZ\cite{10.2307/2632051}  & CSC\cite{DEMIRKOL1998137}   & RC\cite{DEMIRKOL1998137} & SWV\cite{10.2307/2632676}  & TA\cite{TAILLARD1993278}   & YN\cite{inproceedings}   & J60\cite{KOLISCH1997205}  & J90\cite{KOLISCH1997205}  & J120\cite{KOLISCH1997205} \\
            \midrule
                Worse           & 0             & 0                 & 0             & 0             & 0             & 0             & 8             & 3             & 26\\
                Equal           & 0             & 1                 & 2             & 0             & 3             & 0             & 17            & 22            & 91\\
            \midrule
                Better        & 1             & 39                & 8             & 5             & 18            & 3             & 12            & 41            & 170\\
                Closed          & 0             & 0                 & 4             & 0             & 0             & 0             & 0             & 0             & 3\\
            \midrule
                Total           & 1             & 40                & 14            & 5             & 21            & 3             & 37            & 66            & 290\\
            \bottomrule
        \end{tabular}
    }%
    \caption{
        The comparison of the quality of the lower bounds provided by the enhanced \optsol's FDS in the \SI{900}{\second} time limit to the state-of-the-art ones (obtained by any method without any time limit restriction).
        Row "Better" does not contain "Closed" instances.
    }
    \label{tab:sota_comparison}
\end{table}

The columns in \Cref{tab:sota_comparison} represent different groups of benchmark instances.
For each group, rows denote how many times enhanced FDS provided worse, equal or better lower bound than the best found in the literature and how many instances were newly closed by FDS (i.e., the newly found lower bound is equal to the state-of-the-art upper bound, thus representing the value of optimal solution).
The last row contains the total number of instances in the respective benchmark set.
It shows that our enhanced FDS algorithm provides new state-of-the-art results, proving optimality/infeasibility for JSSP and RCPSP problems, yielding hundreds of newly improved lower bounds and a few newly closed instances.
Note that \Cref{tab:sota_comparison} does not account for seven new lower bounds just recently reported in \cite{YURASZECK2025106964}, which, however, required much longer runtimes to be obtained.
Since these results emerged only after our study was completed and submitted, we opted not to update the table, but we acknowledge their contribution.
The lists of all the improved JSSP and RCPSP instances are provided in \ref{app:list_bounds}.
More detailed results are available at \cite{fds_experimental_results}.

\section{Conclusion and Outlook}
\label{sec:conclusion}
Failure-Directed Search (FDS) is a state-of-the-art algorithm designed to efficiently and exhaustively explore the search space of scheduling CCPs.
Its main purpose is to prove the optimality and lower bounds of given (usually but not exclusively) scheduling problem instances.

We showed that FDS is closely related to the Multi-armed bandit (MAB) problem, where rewards encourage the algorithm to minimize the search tree.
We applied existing MAB algorithms (strategies) and introduced a concept of hybrid strategies to improve the performance of the FDS while addressing the well-known exploration-exploitation dilemma.
We also proposed a problem-specific improvement called \emph{choice rollback}, making exploration less costly in the context of FDS.
Finally, parameter tuning was used to adjust both the original FDS parameters and those related to choice-selection strategies.
Since FDS has a structure similar to many search algorithms such as Branch-and-Bound, we believe the presented propositions could be transferred to other problems and algorithms.

The improvements were evaluated on the most fundamental scheduling problems, JSSP and RCPSP.
A hybrid strategy combining greedy strategy with Boltzmann exploration while using \emph{choice rollback} (called \bg$_R$) is \jsspMab\ times faster on JSSP instance set and \rcpspMab\ times faster on RCPSP instance set than FDS without any exploration.
Automatic parameter tuning makes FDS additional \jsspTuning\ times faster on the JSSP instance set and \rcpspTuning\ times faster on the RCPSP instance set.

In conclusion, the enhanced FDS is \jsspToOldFDS\ times faster on JSSP and \rcpspToOldFDS\ times faster on RCPSP benchmarks compared to the original FDS implemention in \optsol.
It is also \jsspToCPOpt\ times faster on JSSP and \rcpspToCPOpt\ times faster on RCPSP than the state-of-the-art FDS implementation in IBM \cpsol\ 22.1.
The enhanced FDS also improves the state-of-the-art lower bounds of \jsspBetter\ out of \jsspAll\ and \rcpspBetter\ out of \rcpspAll\ open JSSP and RCPSP instances, respectively, using just a \SI{900}{\second} time limit, while also closing a few of them.
We believe that these results underline the significance of this research.

In the future, we would like to test our findings on less known, but also interesting instance sets like \cite{COELHO2020104976} that focus on specific problem aspects.
We also aim to evaluate our enhanced FDS on other non-scheduling combinatorial problems, assessing whether the effectiveness of the proposed methods generalizes to different problem domains.
We want to further improve FDS by considering RL with the representation of the solver's internal state and investigate the initial branching choice generation, which was mentioned only briefly in regard to LengthStepRatio and UniformChoiceStep parameters.
Finally, we would like to explore whether a static choice generation could be replaced by a dynamic one that would more effectively adapt to a current variable domain, properties, and relationships to other variables in the current search state.

\section{Acknowledgements}
This work was supported by the Grant Agency of the Czech Republic under Project GACR 22-31670S.
This work was co-funded by the European Union under the project ROBOPROX (reg. no. CZ.02.01.01/00/22\_008/0004590).

\newpage

\appendix

\section{Failure-Directed Search: Algorithm}
\label{app:fds}
FDS is a complete search algorithm that comes from a family of problem-independent dynamic search algorithms such as \emph{Impact-Based Search} (IBS) \cite{refalo2004impact}.
Unlike most search algorithms that navigate towards solutions, FDS tries to prove that no solution exists.
It uses the modified \FF\ principle, which means that it tries to make decisions on variables that will likely lead to producing a proof of infeasibility.
This is straightforward for CSPs as FDS either proves the whole search space is infeasible or, while doing that, finds a feasible domain-variable assignment (solution).
But for COPs, FDS is designed to be \pB, complementing other \pA\ search strategy.
In our solver \optsol, this \pA\ strategy is Large Neighborhood Search (LNS) \cite{shaw1998using, godard2005randomized}, but it can be any strategy that provides quality (good criterion value) solutions quickly without requiring optimality guarantee. 
FDS then uses the provided solution's criterion value as a bound, trying to prove that no feasible solution with a better criterion value exists, ensuring its optimality.
The better the solution provided, the faster FDS explores the whole solution space.
If no solution is provided by \pA\ strategy, FDS still explores the search tree and closes infeasible branches, but less efficiently without the additional constraint on criterion value.
If FDS finds a new feasible solution as a byproduct of its search, the solution is also used to provide a bound.

When FDS is executed, it explores the search tree in a depth-first manner.
Each child node of the tree represents a smaller solution space than its parent because, at each node, a domain of some variable is reduced.
This reduction is achieved by adding new constraints given by the so-called \emph{choices} and then using \emph{constraint propagation} in each node: the constraints are taken one by one, and a constraint-specific algorithm removes infeasible values from the variable domains or detects infeasibility (the so-called \texttt{Fail}).
Some of the used constraint propagation algorithms in \optsol\ are: Timetable filtering and Edge finding \cite{edge_finding_vilim, 10.1007/978-3-319-23219-5_11, 10.1007/978-3-642-40627-0_42}, Detectable precedences \cite{vilim_disertace}, Not-first/Not-last constraints \cite{vilim_disertace}, and Disjunctive constraint \cite{baptiste2001}.
Because constraint propagation can detect infeasibility even on partially reduced domains, FDS tries to order the choices so the infeasibility is detected as soon as possible (the \FF\ principle), avoiding exploration of all possible nodes.
Thus, each tree leaf node contains either a solution or proof that there is no feasible assignment of values to variables within the current domains.

\subsection{Choices and Branches}
A \emph{choice} is a representation of a binary decision on a variable that constrains its domain to two disjunct intervals (branches).
An example of a choice for some variable \varm\ would be "$\var = 5$ OR $\var \ne 5$".
In the case of FDS, domain splitting is used to make the number of choices reasonable, so every choice splits the domain into two continuous intervals.
An example of such a choice would be "$\var \le 5$ OR $\var > 5$".
Thus, the choice can be represented as a tuple (variable, pivot), splitting the variable domain into two possible branches in the pivot's value.

At the start of the algorithm, a set of choices for every variable is generated (explained later) based on parameters like granularity of choices, etc.
This is done by calculating how many times (and where) the domain will be split, and for each of these points, the choice that splits the domain into two parts is created.
FDS itself does not consider the choice's effect; it only knows that the choice creates two branches, left and right.
The algorithm explores the left branch first, but if the rating (explained later) of the right branch becomes better than the rating of the left branch, the left branch becomes the right branch, and vice versa.
The outcome of \emph{constraint propagation} for the new branch (whether infeasibility was detected or domains were reduced) is passed to the FDS.

The order in which FDS adds the choices to the search tree is very important.
There are many heuristics such as \emph{dom/wdeg} \cite{boussemart2004boosting}, \emph{wged} \cite{8995307}, and \emph{CHS} \cite{habet:hal-02090610} that are used to determine the variable priority of the search algorithm.
In FDS, the order of the choices is given by their ratings.
Ratings are repeatedly updated and denote the strength of a particular choice's propagation.
Each choice $c$ has two ratings: the rating of its left branch ($\rpl$) and the right branch ($\rmi$).
The initial rating value is the same for both branches (FDS parameter \dr).
Choice's rating, denoted $\ra$, is then:
\begin{gather}
  \ra := \rpl + \rmi.
  \label{def:rating}
\end{gather}
In each step, FDS selects the choice with the best $\ra$ (smaller means better).
The ties are broken randomly, which is important, especially at the beginning of the search (all ratings are the same).
The result of the constraint propagation is represented by $\lr$ calculated as follows:
\begin{gather}
   \lr :=
   \begin{cases}
      0  &\text{\rmfamily if infeasibility is detected}, \\
      1 + R &\text{\rmfamily otherwise}.
   \end{cases}
  \label{def:localRating}
\end{gather}

The $R$ represents a reduction factor calculated as $0.5^{n}$ where $n$ denotes the number of variable domains reduced by the constraint propagation in this node.
Notice that $\lr$ is a number from the set $\{0\}\cup(1, 2]$.
Subsequently, the rating of the branch ($\rpl$ or $\rmi$) is updated using the following formula:
\begin{gather}
  \rb := (1-\alpha) \cdot \rb + \alpha \cdot \lr,
  \label{def:ratingupd}
\end{gather}
where $\alpha$ is a parameter from the range $[0,1]$ controlling the speed of rating change
\footnote{
  Remark: In the original FDS \cite{vilim2015failure}, $\lr$ was multiplied by $1-\alpha$.
  To follow the usual convention, we changed it to $\alpha$ in \Cref{def:ratingupd} (the value was adjusted accordingly to maintain the same effect).
  Also note that the average rating on the current depth $d$ ($\avd$) used in the original FDS is no longer present in \Cref{def:ratingupd}.
  More details on the removal of $\avd$ are provided in \ref{app:avd}, as they are not essential to understand the algorithm.
  }.
Notice that the left and right $\alpha$ can have different values, which is explained in more detail in \ref{subsec:factor_alpha}.

\begin{figure}[ht]
    \centering
    \scalebox{0.88}{
    \includegraphics[width=0.6\linewidth]{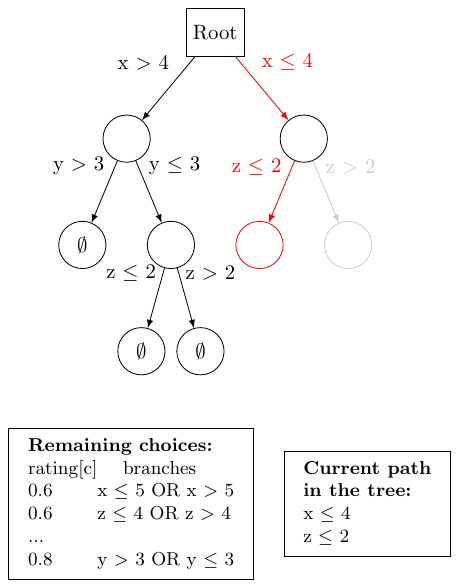}
    }
    \caption{
        Red color denotes the current path.
        The "Remaining choices" are ordered by their ratings (ties are broken randomly).
    }
    \label{img:branching}
\end{figure}

The resulting $\rb$ is always a number in the interval $[0,2]$.
An example of a search tree branching on choices is in \Cref{img:branching}.

Sometimes, the best-rated choice is already \texttt{Decided}: one of the branches is true, and the other is false.
Consider the situation in the search tree in \Cref{img:branching}, representing the branching with $\dr=0.3$.
The domain of \varm\ is $\{1,2,3,4\}$ due to the previous branching in the tree and by the propagation in the nodes.
Now, the best-rated choice is $\var \le 5$ OR $\var > 5$.
Clearly, the branch $\var \le 5$ is already true and $\var > 5$ is false, making it useless to branch on this choice.
Thus, the choice $\var \le 5$ OR $\var > 5$ is put aside (reversibly, so it can be picked after backtracking when the domain of \varm\ is larger again), and the next best-undecided choice (e.g., $z \le 2$ OR $z > 2$) is picked instead.
Please note that in this paper we only consider (non-optional) interval variables.

FDS employs a geometric restart strategy \cite{gomes2000heavy, moskewicz2001chaff}.
The restart of the search is a process in which the search tree is rebuilt from scratch, but the ratings of the choices (branches) are kept.
The idea is that with each restart, FDS gets better at the choice ordering, finding \emph{Fails} earlier in the search tree, thus reducing its size.
The number of branches explored before a restart is triggered is initially given (FDS parameter FDSInitialRestartLimit) and increases geometrically after each restart (FDS parameter FDSRestartGrowthFactor).
FDS also uses no-good recording from restarts \cite{lecoutre2007nogood}: at the point of the restart, new constraints are added into the system in order to prevent the solver from exploring the same search space again.
The no-good constraints are constructed from the current path in the search tree at the time of restart.

\subsection{Flowchart and Pseudocode}
\begin{figure}[ht]
    \centering
    \includegraphics[width=\linewidth]{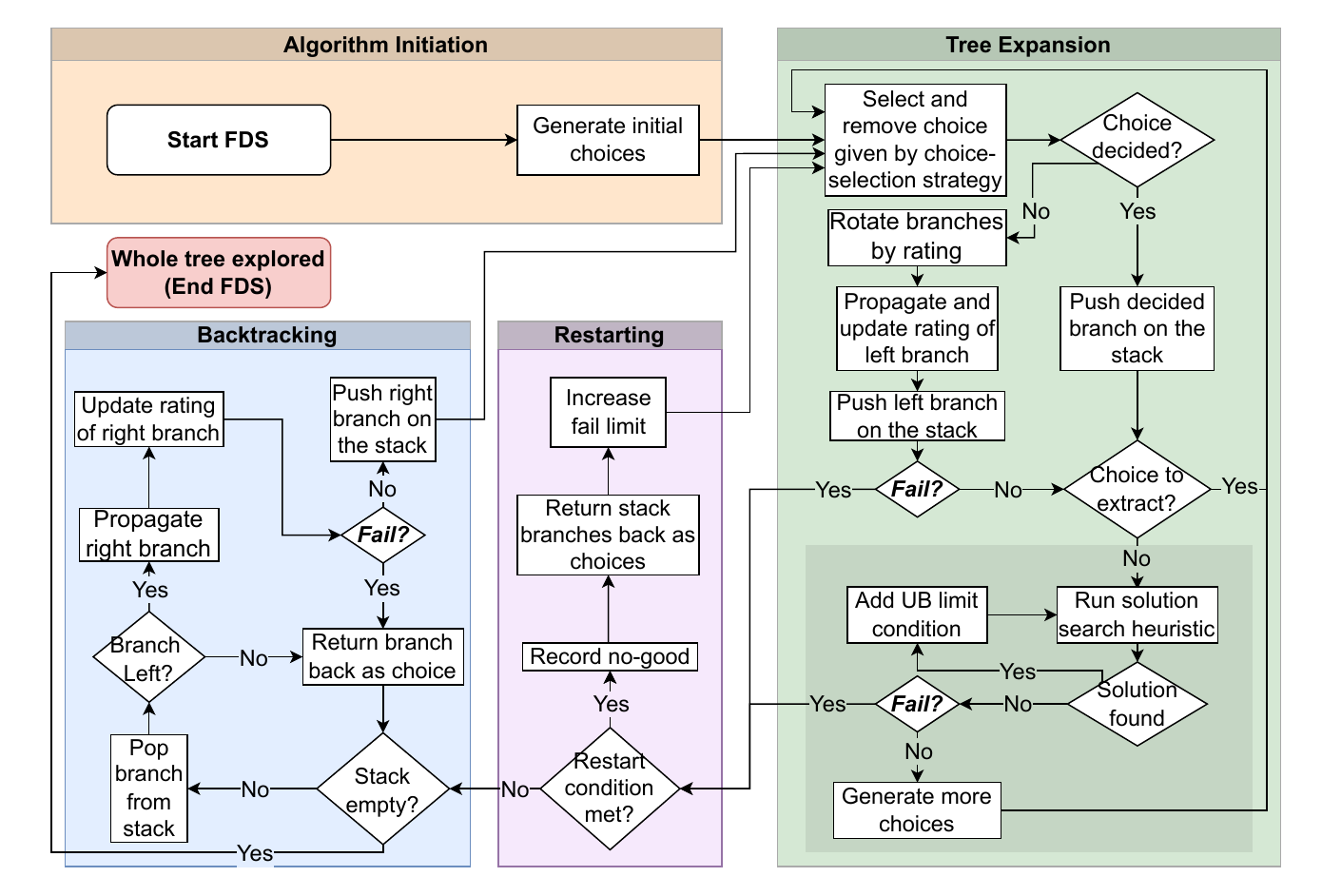}
    \caption{Flow chart of the FDS algorithm.}
    \label{img:fds}
\end{figure}

Since the explanation for the FDS algorithm enclosed in the original paper \cite{vilim2015failure} is not sufficiently detailed, we provide our own pseudocode in \Cref{alg:fds} and a flow chart in \Cref{img:fds} to give a detailed and unambiguous description of the algorithm.
Note that both are using shared color-coding to highlight and distinguish different logical parts of the algorithm.
Additional functions used in the pseudocode are explained in \Cref{tab:fds_fnc}.
For brevity, in the following text, we define $\C$ as the complete set of problem constraints.

\begin{singlespace}
\begin{figure}
    \centering
\noindent
\scalebox{0.9}{
\begin{minipage}{1.1\textwidth}

    \begin{algorithm}[H]
    \DontPrintSemicolon
    \scriptsize
    \caption{Pseudocode of the FDS algorithm.}
    \label{alg:fds}

    \KwData{$\fl, \rf, \C$}

    \hbox{\hspace{-0.36em}
        \begin{tcolorbox}[
                    colback=orange!12!white,colframe=green!10!white, 
                    left=0mm,right=0mm, top=0mm, bottom=0mm, 
                    boxsep=0mm,boxrule=0mm, width=142mm, 
                    after=\vspace{-0.2\baselineskip},
                    before=\vspace{-0.2\baselineskip},
                ]
            \tcc{Algorithm Initiation}
            $\fc \gets 0 $ \tcp*{set fail count} 
            \label{code:beginInit}
            $\st \gets$ New Stack \tcp*{(empty) stack structure representing selected branches}
            $\hp \gets \GIH{} $    \tcp*{choices generated from variable domains}
            \label{code:endInit}
        \end{tcolorbox}
    }

    \While{$True$}{
    
        \hbox{\hspace{-0.36em}\vspace{0.2em}
            \begin{tcolorbox}[
                colback=green!11!white,colframe=green!10!white, 
                left=0mm,right=0mm, top=0mm, bottom=0mm, 
                boxsep=0mm, boxrule=0mm, width=135mm, 
                after=\vspace{-0.2\baselineskip},
                before=\vspace{-0.2\baselineskip},
            ]
                \tcc{Tree expansion}
                \While {$\hp \ne \{\}$}{
                \label{code:beginTree}
                    $\bc \gets \SC{\hp}$ \tcp*{get next choice by the selection criteria}
                    \label{code:beginDecided}
                    $\hp \gets \hp \setminus \{\bc\}$ \;

                    \If{\ID{$\C$, \st, \bc}}{
                        \st.\ps{\bc.\tb} \tcp*{insert element into stack}
                        \label{code:endDecided}
                    }
                    \Else{
                    \label{code:beginLeft}
                        \RB{\bc} \tcp*{ensure that branch with lower rating is left}
                        \tcc{propagation returns \emph{\lr} depending on branch stack (with new branch)}
                        $\emph{\lr} =\ $\PC{$\C$, \st, \bc.\lbr} \;
                        $\UBR{\bc.\lbr, \lr}$  \tcp*{updates rating of given branch}
                        \st.\ps{\bc.\lbr} \tcp*{insert element into stack}
                        \If{\lr\ = 0}{
                            \textbf{break} \;
                            \label{code:endLeft}
                        }
                    }

                    \tcc{if we branched all available choices and solution space is still feasible}
                    \If{$\hp = \{\}$}{
                    \label{code:beginEmpty}
                        \While(\tcp*[f]{while solution space is feasible}){\PC{$\C$, \st, $True$} $\ne 0$} {
                            \sol = \FFA{$\C$, \st} \tcp*{heuristic search}  
                            \If{\sol = NULL}{
                                \tcc{divide remaining domains into smaller branching intervals} 
                                $\hp \gets $ \GMC{$\C$, \st} \;
                                \label{code:newChoices}
                                \textbf{break}
                            }
                            $\C \gets \C \cup \{\LO{\sol.Value - 1}\}$ \tcp*{reduce UB for new solutions}
                            \label{code:endTree}
                        }
                    }
                }
            \end{tcolorbox}
        }

        $\fc \gets \fc + 1$ \; 

        \hbox{\hspace{-0.24em}\vspace{0.6em}
        \begin{tcolorbox}[
                colback=blue!10!white,colframe=green!10!white, 
                left=0mm,right=0mm, top=0mm, bottom=0mm, 
                boxsep=0mm,boxrule=0mm, width=135mm, 
                after=\vspace{-0.2\baselineskip},
                before=\vspace{-0.2\baselineskip},
            ]
            \tcc{Backtracking}
            \While{$\fc \ne \fl$}{
            \label{code:beginBacktracking}
                \If{$\st = \{\}$}{
                \label{code:checkexit}
                    \colorbox[rgb]{1.0,0.65,0.65}{\textbf{exit} \tcp*{whole search space was explored}}
                    \label{code:exit}
                }
                $\br \gets \st.\pp$ \tcp*{extract recent branch}

                \If(\tcp*[f]{expand right side of branch's choice}){\CIL{\br}}{
                \label{code:isLeft}
                    $\bc = \GC{\br}$ \tcp*{get parent choice of branch}
                    \label{code:getChoice1}
                    $\emph{\lr} =\ $\PC{$\C$, \st, \bc.\rbr} \;
                    \UBR{\bc.\rbr, \lr} \;
                    \tcc{The left branch is already explored, check if the right branch failed}
                    \If{$\lr > 0$}{
                    \label{code:beginPushRight}
                        \st.\ps{\bc.\rbr} \tcp*{insert element into stack}
                        \textbf{break} \tcp*{returns to tree expansion}
                        \label{code:endPushRight}
                    }
                    $\fc \gets \fc + 1$ \; 
                }
                
                $\hp \gets \hp \cup \{\GC{\br}\}$ \tcp*{add choice back}
                \label{code:endBacktracking}
            }
        \end{tcolorbox}
        }

        \hbox{\hspace{-0.24em}\vspace{0.6em}
        \begin{tcolorbox}[
                colback=purple!12!white,colframe=green!10!white, 
                left=0mm,right=0mm, top=0mm, bottom=0mm, 
                boxsep=0mm,boxrule=0mm, width=135mm, 
                after=\vspace{-0.2\baselineskip},
                before=\vspace{-0.2\baselineskip},
            ]
            \tcc{Restarting}
            \If{$\fc = \fl$}{
            \label{code:beginRestarting}
                \tcc{add current branches' stack as no-good}
                $\C \gets \C \cup \GNGC(\st)$ \;
                \label{code:noGood}
                \tcc{add branches back as the original choices}
                \ForEach{$\br \in \st$}{
                \label{code:startClear}
                    $\hp \gets \hp \cup \{\GC{\br}\}$ \tcp*{add branch's choice back} 
                }
                $\st \gets \{\}$ \tcp*{reset stack}
                \label{code:endClear}
                $\fl \gets \fl * \rf$ \;
                \label{code:failCountIncr}
                $\fc \gets 0$ \\
                \label{code:endRestarting}
            }
        \end{tcolorbox}
        }
    }
    \end{algorithm}
\end{minipage}
}
\end{figure}

\begin{table}
    \centering
    \begin{description}[itemsep=0.25em]

        \item{\GIH{}} returns a set of initial \hp.
        Since generation of initial choices is beyond this paper's scope, we provide its detailed explanation in \ref{app:gic}.
        
        \item{\SC{\hp}} returns \bc\ with the best rating from \hp. Please note that, in \Cref{sec:choice_selection}, this function is extended using RL MAB algorithms.
        
        \item{\ID{$\C$, \st, \bc}} returns \textit{true} if after propagation of constraints in $\C$ and \st, the given \bc is \emph{Decided}, meaning one of its branches is clearly \textit{true} and the other \textit{false}.
    
        \item{\RB{$\bc$}} ensures that branch with lower rating will be on the left by switching the branches' positions if necessary.
        
        \item{\PC{$\C$, \st, \br}} propagates constraints from $\C$, the current search decisions from $\st$, and the new search decision \br.
        The function returns \emph{\lr}\ of the \br, as defined by \Cref{def:localRating}.
        In particular, the function returns 0 if an infeasibility (\emph{Fail}) is detected.
        In practice, the propagation starts from variable domains computed already in the parent node.
        
        \item{\UBR{\br, \lr}} updates the rating of the branch (one side of the choice) according to the formula in \Cref{def:ratingupd}.
        
        \item{\FFA{$\C$, \st}} heuristically tries to solve the problem with constraints $\C$ and \st. In our case, the used heuristic is {\em SetTimes} strategy from \cite{godard2005randomized} without backtracking.
        
        \item{\GMC{$\C$, \st}} returns a set of additional choices that split current variable domains after propagation of \st and $\C$ into finer intervals (choice again splits the domain into two intervals).
        The remaining domain of every variable is split evenly into maximum of 5 additional choices.
        
        \item{\LO{\ub}} returns a new constraint that limits the objective value not to exceed \ub.
    
        \item{\CIL{\br}} returns \textit{true} if current \br is a left branch, \textit{false} otherwise.
    
        \item{\GC{\br}} returns \bc associated with the given \br.
        
        \item{\GNGC{\st}} returns a set of new nogood constraints generated from the current path given by \st. The algorithm used is described in \cite{lecoutre2007nogood}.
        
    \end{description}
    \caption{Functions used in \Cref{alg:fds}.}
    \label{tab:fds_fnc}
\end{table}
\newpage
\end{singlespace}

\textbf{\paragraph{Initiation}}
\textbf{Initiation (Lines \ref{code:beginInit} to \ref{code:endInit}): }
Initiation contains the creation of a counter \fc to calculate when the search should be restarted, a stack structure \st for selected branches representing the current path in the search tree, and a structure \hp holding the generated (at the start or in the runtime) choices (\GIH).

\textbf{Tree expansion (Lines \ref{code:beginTree} to \ref{code:endTree}): }
Tree expansion is where the branches are added to \st.
A \bc is selected by a given criterion (\SC) and checked if it is not \texttt{Decided}, see lines \ref{code:beginDecided} to \ref{code:endDecided}.
If \texttt{Decided}, the selection of choice is repeated.
If not, the branches of the choice are rotated so that the one with a lower rating (more likely to fail) is on the left.
Then, the propagation of all constraints is executed (\PC), the rating of the branch is updated (\UBR), and if \texttt{Fail} ($\lr = 0$) is reached, backtracking or restarting follows, see lines \ref{code:beginLeft} to \ref{code:endLeft}.
In case the algorithm runs out of \hp, a solution is repeatedly searched for (using a greedy search strategy) (\FFA), which provides an increasingly tight upper bound cut (\LO), see lines \ref{code:beginEmpty} to \ref{code:endTree}.
If no solution is found, but solution space still could be feasible (meaning \PC on line 17 did not fail), new choices must be generated on line \ref{code:newChoices} (\GMC).
The new choices split the resulting variable domains into finer intervals so that the algorithm may continue branching.
It is important to note that once the algorithm moves from tree expansion to branching/restarting (line \ref{code:beginBacktracking}), the currently explored branch is infeasible.

\textbf{Backtracking (Lines \ref{code:beginBacktracking} to \ref{code:endBacktracking}): }
The backtracking process is repeated until one of three things occur.
One, \st becomes empty, which means the solution space is infeasible, and the search ends; see lines \ref{code:checkexit} and \ref{code:exit}.
Two, the \fc is equal to \fl, meaning the algorithm should be restarted; see line \ref{code:beginBacktracking}.
Third, the right branch of the parent choice retrieved on line \ref{code:getChoice1} (\GC) is still feasible, and the algorithm returns to tree expansion; see lines \ref{code:beginPushRight} to \ref{code:endPushRight}.
The third option can only happen if the branch retrieved from the stack is left (\CIL); see line \ref{code:isLeft}.
If the retrieved choice was right, it would mean that both branches were already explored, so \bc is returned to \hp.
In such case, the \bc is returned to \hp, see line \ref{code:endBacktracking}.
This also happens if the newly explored right branch is infeasible.

\textbf{Restarting (Lines \ref{code:beginRestarting} to \ref{code:endRestarting}) :}
Restarting is only triggered if \texttt{Fail} occurs somewhere in the code and the \fc is equal to \fl.
When the algorithm is restarted, a no-good is generated (\GNGC), ensuring that new searches do not repeat exploration of the same subspace; see line \ref{code:noGood}. 
Furthermore, all choices from \st are returned to \hp, and \st is cleared; see lines \ref{code:startClear} and \ref{code:endClear}.
The result is essentially a reset of the search but with updated ratings and additional no-goods.
Finally, \fl is then increased, and \fc is reset; see lines \ref{code:failCountIncr} and \ref{code:endRestarting}.

\subsection{Strong Branching}
\label{subsec:strong_branching}
Strong branching, initially introduced in \cite{10.5555/868329} and explained in detail in \cite{achterberg2005branching}, is used to fine-tune choice selection in FDS.
Instead of picking one choice and then branching on it, the idea is to evaluate a certain number of the best available choices (FDS parameter StrongBranchingSize) without adding them to the search tree.
For each of them, constraint propagation is run, but a backtrack is done immediately after each one to return to the original state.
Then, the choice that propagated the most is used for real branching.
The propagation is either evaluated only by the rating of the left branch, by the rating of the right branch, or by the rating of both branches together (FDS parameter StrongBranchingCriterion).
However, strong branching is quite costly, so in addition to limiting it by the number of choices tested, strong branching is also used only to a limited depth in the search tree (FDS parameter StrongBranchingDepth).
In pseudocode, strong branching is part of the logic of the function \SC on line \ref{code:beginDecided}.
Instead of simply selecting the choice with the best rating, multiple choices would be selected to evaluate the strength of their propagation if added to the search tree, and only the one with the largest propagation would be permanently added to the tree (that is, returned by the function \SC as the choice).

\subsection{Learning Rate \texorpdfstring{$\alpha$}{Alpha}}
\label{subsec:factor_alpha}
While IBS \cite{refalo2004impact}, which inspired FDS, originally used arithmetic average to compute impacts (an analogy to FDS ratings), the original FDS uses static $\alpha$ instead.
The main advantage compared to the arithmetic average is that $\rb$ can quickly adjust to changes in $\lr$ even after many updates.
This is important because ancestor nodes in the search tree might heavily influence current $\lr$.
However, using static $\alpha$ instead of the arithmetic average means
that $\rb$ moves more slowly from its initial value at the beginning of the search.
To balance the two, a sort of hybrid between $\alpha$ and arithmetic average is proposed.

In particular, for each choice $c$, two counters are maintained: $\cle$ and $\cri$ for the number of times $\rpl$ and $\rmi$ were updated, and they are increased until they reach value~$L$ (FDS parameter RatingAverageLength).
So, this "hybrid $\alpha$" of the choice used in \Cref{def:ratingupd} is calculated in the following way:
\begin{align}
  \clr &:= \min\left\{ \clr + 1, L  \right\}, \\
  \alpha &:= 1 / \clr.
  \label{def:alpha}
\end{align}
Thus, the first $L$ updates of $\ra$ in the \Cref{def:ratingupd} are equivalent to the arithmetic average of the obtained ratings for the choice so far.
After $L$, the factor $\alpha$ remains constant ($\alpha = 1/L$).
For syntactic clarity and consistency with the notation used in \cite{vilim2015failure}, the symbol $\alpha$ in \Cref{def:alpha} is used without explicitly distinguishing between the left and right branches.
However, it is important to note that its value may differ depending on the branch.

\section{Failure-Directed Search: Supplementary}

\subsection{Description of FDS Parameters}
\label{app:fds_params}
\begin{itemize}[noitemsep,topsep=0pt,parsep=0pt,partopsep=0pt]
    \item \textbf{AdditionalStepRatio}: How finely domains should be split if all choices were decided and no solution was still found.
    \item \textbf{BothFailRewardFactor}: How much to further improve choice rating if both branches failed immediately.
    \item \textbf{InitialRating}: Initial rating value for newly created choices.
    \item \textbf{Epsilon}: How often select the exploratory choice.
    \item \textbf{EpsilonDecay}: How fast the epsilon is decreased.
    \item \textbf{EventTimeInfluence}: How much is the initial choice rating influenced by its relative chronological order. 
    \item \textbf{FixedAlpha}: Static alpha factor value.
    \item \textbf{InitialRestartLimit}: How many fails before first restart.
    \item \textbf{LengthStepRatio}: Choice step relative size to the average length.
    \item \textbf{MaxCounterAfterRestart}: Maximum value of variable use counter $\clr$ after a restart.
    \item \textbf{MaxCounterAfterSolution}: Maximum value of variable use counter after finding a solution.
    \item \textbf{MaxInitialChoicesPerVariable}: Max initial choice count per variable.
    \item \textbf{PenaltyNotFailed}: \textit{No longer supported by \optsol.}
    \item \textbf{PresenceStatusChoices}: If consider presence status in choice generation.
    \item \textbf{RatingAverageComparison}: If ratings are relative to average.
    \item \textbf{RatingAverageLength}: Number of ratings considered in average $L$.
    \item \textbf{ResetRestartsAfterSolution}: Reset restart size after a solution is found.
    \item \textbf{RestartGrowthFactor}: Growth factor for the fail limit after each restart.
    \item \textbf{RestartStrategy}: Restart strategy to use (geometric, nested or Luby).
    \item \textbf{StartVariableAlphaN}: \textit{No longer supported by \optsol.}
    \item \textbf{StrongBranchingCriterion}: How to choose the best choice from the tested ones in strong branching.
    \item \textbf{StrongBranchingDepth}: Max search depth to apply strong branching.
    \item \textbf{StrongBranchingSize}: Number of choices to try in strong branching.
    \item \textbf{UseNogoods}: Whether to use nogood constraints or not.
\end{itemize}
More details on all parameters are provided in \cite{optalcp_params}.

\subsection{Removal of \texorpdfstring{$\avd$}{Average Rating}}
\label{app:avd}
The original version of \Cref{def:ratingupd} that calculated the rating of a branch took into account the average choice rating on the current tree depth $d$ (denoted $\avd$).
It looked as folows (please note that in \cite{vilim2015failure}, branches were denoted by $\pm$ instead of left and right):
\begin{gather}
  \rbx := (1-\alpha) \cdot \rbx + \alpha \cdot \frac{\lr}{\avd}.
  \label{def:origrating}
\end{gather}

The original FDS paper \cite{vilim2015failure} explains the importance of $\avd$ by necessity to take into account different amounts of average constraint propagation on different search depths.
The argument is that on the first levels of the tree, there will be almost no propagation (domain reduction), while on deeper levels, selected branches will propagate much more, so it should be reflected by rating calculation.
However, as it turns out, this idea of different average propagation on each level of the tree is not the reason why $\avd$ was initially efficient.

\begin{figure}[htb]
    \centering
    \includegraphics[width=\linewidth]{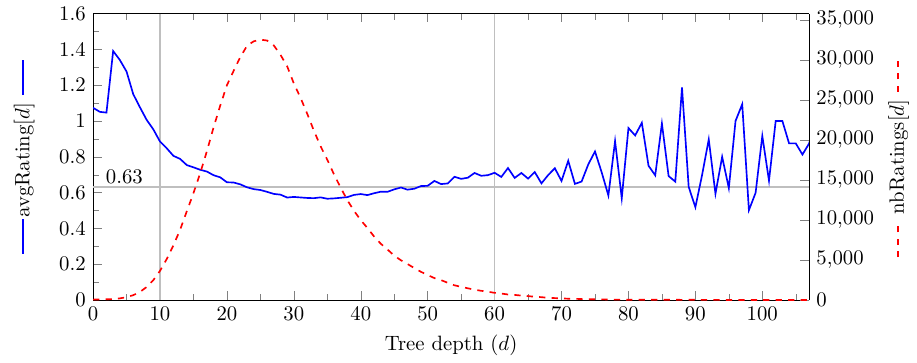}
    \caption{$\avd$ on different depths $d$ of the tree. The X-axis denotes the depth. The left Y-axis denotes the $\avd$ while the right Y-axis shows the number of ratings on the specific depth.}
    \label{img:avg}
\end{figure}

Let us examine the effect of $\avd$ on a JSSP instance \texttt{ta50} with an additional constraint forcing the makespan to be $1832$ or less, making it infeasible.
FDS was run 100 times with different random seeds (as ties are broken randomly), exploring the whole search space and proving infeasibility while measuring $\avd$ on different tree depths shown in \Cref{img:avg}.

The obtained data show that $\avd$ actually does not change that much.
The global average of $\avd$ (calculated over all depths) is approximately $0.63$, and the deviation is not very pronounced.
The overall execution time can be seen on row 1 of \Cref{tab:avg_rating}.
Row 2 of \Cref{tab:avg_rating} shows that when we use this knowledge and replace $\avd$ with a constant equal to this average, it even leads to a slight improvement.
So, it might seem that the use of $\avd$ was redundant.
However, the result on row 3 of \Cref{tab:avg_rating} shows that when $\avd$ is instead replaced with a constant $1$ (basically omitted), the instance cannot be solved within one hour.
This is because changing $\avd$ from $0.63$ to $1$ effectively multiplies all $\lr$ in \Cref{def:origrating} by $0.63$, making $\lr$ that much smaller.
This means that every update of $\ra$ with $\avd=1$ makes the choice more preferable than it would with $\avd=0.63$ and more preferable against the non-updated choices.

\begin{table}[htb]
\footnotesize
\centering
    \includegraphics[width=\linewidth]{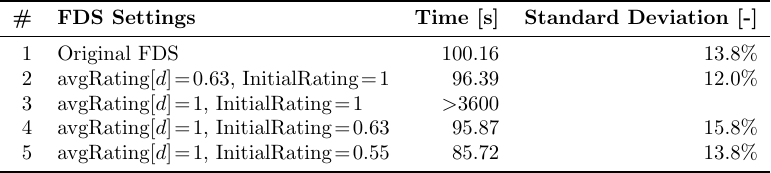}
    \caption{Evaluation of $\av$ and $\dr$ effect on the original FDS algorithm measured on JSSP instance \texttt{ta50} with maximum allowed makespan $1832$.}
    \label{tab:avg_rating}
\end{table}

Thus, $\avd$ can be effectively replaced by introducing a new, more straightforward parameter called \dr, which is a reason why we no longer consider $\avd$.
As the name suggests, this parameter sets the initial rating value of every choice, which can be used to substitute the effect of $\avd$ in a more straightforward way.
Please note that if both $\avd$ and $\dr$ were used, it would lead to effectively multiplying the rating twice and a worse result.
The solution is to adjust $\dr$ to accommodate for the omitting of $\avd$.
Row 4 of \Cref{tab:avg_rating} shows that with $\dr=0.63$, the performance is again on par with row 2.
If $\dr=0.55$ is used instead, the performance is even better; see row 5 of \Cref{tab:avg_rating}.
As it turns out, $\dr$ is an influential parameter, and its tuning was discussed in \Cref{sec:param_tuning}.

\subsection{Generation of Initial Choices}
\label{app:gic}
The function generating the initial choices (\GIH) first computes a so-called \text{Step}, i.e., a distance between two neighboring splitting points (pivots) in the variables' domains.
Depending on the UniformChoiceStep parameter, the Step size calculation uses the average of all intervals (true) or not (false):
\[
    \text{Step} = 
    \text{LengthStepRatio} \cdot
    \frac{\sum_{v \in \V} \text{lengthMin}(v)}{|\V|}; \quad \quad \text{(UniformChoiceStep=1)}
\]

\vspace{-4em}
\[
    \text{Step} = 
    \text{LengthStepRatio} \cdot \text{lengthMin}(v); \quad \quad  \text{(UniformChoiceStep=0)}
\]

In both equations, LengthStepRatio is a parameter indicating the scale of the Step (its default value is $0.7$), and lengthMin is a function that returns the minimum length in the domain of the interval variable $v$ (in the case of JSSP and RCPSP, the length of the specific interval variable is always the same).
$\V$ denotes the set of all interval variables.
Thus, $|\V|$ denotes the total number of them.

In the next step, the pivots, denoted by $\Pe$, are calculated:
\begin{multline*}
  \Pe_v = \biggl\{ 
    \text{round}\left(\text{startMin}\left(v\right) + i \cdot \text{Step} \right)
    ,\;\\
    i = 1 \dots \left\lfloor \frac{\text{startMax}(v) - \text{startMin}(v)}{\text{Step}} \right\rfloor
  \biggr\} \;\;\;\; \forall v \in \V
  .
\end{multline*}

The functions startMin and startMax provide the minimum and maximum start time in the initial domain of the variable $v$.
Finally, \hp\ are generated as:
\[
    \hp = \left\{
        \text{startOf}\left(v\right) \leq p \lor \text{startOf}\left(v\right) > p, \; 
        \forall v \in \V, \; \forall p \in \Pe_v
    \right\} 
\]

The initial rating of all branches of all choices is given by the parameter \dr.

\subsection{Exploration Effect}
\label{app:fixrat}
\begin{figure}
    \centering
    \scalebox{0.8}{
        \begin{minipage}{0.6\textwidth}
        \includegraphics[width=\linewidth]{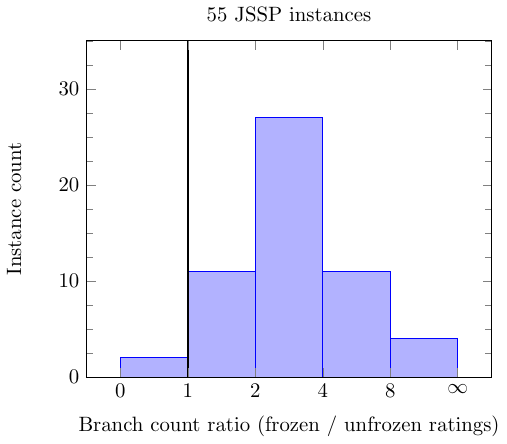}
        \end{minipage}
        \hfill
        \begin{minipage}{0.6\textwidth}
        \includegraphics[width=\linewidth]{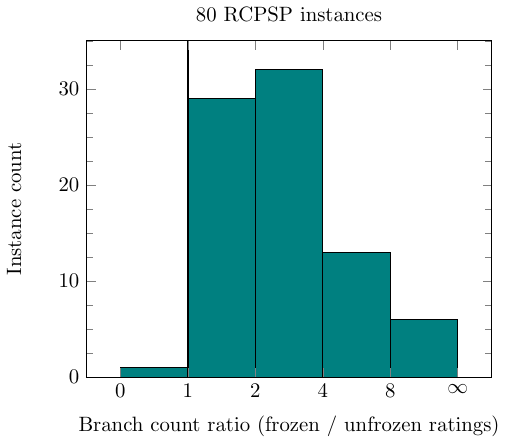}
        \end{minipage}
    }

    \caption{
        Histograms show the number of instances that have the ratio of explored branches when comparing runs with and without rating freeze in particular intervals.
        Each bin represents doubled runtime of the previous bin with the exception of the last bin that contains all remaining instances.
        The ratings were initially trained and carried over from 10 previous runs of FDS on the same instance.
        Thus, a bin between 0 and 1 means that freezing the learned ratings after 10 complete runs was beneficial, while all other bins represent the instances where it was better to keep ratings unfrozen.
    }
    \label{fig:histogram}
\end{figure}

To demonstrate that it is beneficial not to fix the ratings (stop exploring) even though they have been learned very well, we provide the following experiment.
In \Cref{fig:histogram}, histograms showing the ratio between two different types of runs are provided.
One type of run reuses the ratings learned from 10 previous complete FDS executions while freezing them in the next run.
The other also reuses the same learned ratings but does not freeze them.
As the ratios from the experiment demonstrate clearly, the majority of the instances are worse off when the ratings are fixed to learned values, even if they have been learned over the course of 10 complete runs.
The average runtime per instance with fixed ratings is 7.5 times larger for 55 tested small and mid-sized JSSP instances and 3.8 times larger for 80 small and mid-sized RCPSP instances than if the ratings are not fixed; same instances as in \Cref{subsec:effects_learning} were used.
While it is hard to understand exactly why, we believe that the main reason is the context (all current constraints in the problem) in which the branch is evaluated.
In other words, the rating of the branch also depends on the complete set of constraints that are currently present, and it might be beneficial to adjust the rating accordingly.

\section{Results: Statistical Analysis}
\label{app:stat_anal}
To assess the statistical significance of differences between MAB algorithms and selected configurations in Tables \ref{tab:mab_comparison} and \ref{tab:all_comparison} respectively, we performed pairwise Wilcoxon signed rank tests and paired t-tests on all test instances, using mean performance across seeds for each algorithm.
The tables provided below show the mean difference values per instance for all pairs of algorithms.
If value in a cell is negative, it means that algorithm/configuration given on a row is that much faster (in seconds) on average per instance of the dataset than the algorithm given in the column.
Differences that are found to be statistically significant using the level of p $<$ 0.05 are indicated in bold, and those that are found statistically significant using the level of p $<$ 0.005 are also denoted by a star.

\subsection{MAB Algorithms}
The results of both Wilcoxon and paired t-test in Tables \ref{tab:map_mab_paired_JSSP}, \ref{tab:map_mab_paired_RCPSP}, \ref{tab:map_mab_wilcoxon_JSSP}, and \ref{tab:map_mab_wilcoxon_RCPSP} demonstrate that with the exception of JSSP instances, the \bgHR\ algorithm is proven to be better than any other configuration with strong statistical significance.

\begin{table}[ht]
    \centering
    \begin{tabularx}{\linewidth}{l|*6{X}}
        \toprule
         & \textbf{\tgL} & \textbf{\tgLR} & \textbf{\egL} & \textbf{\egHR} & \textbf{\bgL} & \textbf{\bgHR} \\
        \midrule
        \textbf{\tgL} & -- &  &  &  &  &  \\
        \textbf{\tgLR} & \cellcolor{white} \textbf{-24.41$^{*}$} & -- &  &  &  &  \\
        \textbf{\egL} & \cellcolor{white} \textbf{-8.81} & \cellcolor{white} \textbf{15.61$^{*}$} & -- &  &  &  \\
        \textbf{\egHR} & \cellcolor{white} \textbf{-23.14$^{*}$} & \cellcolor{white} \textbf{1.27} & \cellcolor{white} \textbf{-14.34$^{*}$} & -- &  &  \\
        \textbf{\bgL} & \cellcolor{white} -14.67 & \cellcolor{white} \textbf{9.74$^{*}$} & \cellcolor{white} \textbf{-5.87$^{*}$} & \cellcolor{white} \textbf{8.47$^{*}$} & -- &  \\
        \textbf{\bgHR} & \cellcolor{white} \textbf{-26.36$^{*}$} & \cellcolor{white} \textbf{-1.95$^{*}$} & \cellcolor{white} \textbf{-17.56$^{*}$} & \cellcolor{white} -3.22 & \cellcolor{white} \textbf{-11.69$^{*}$} & -- \\
        \bottomrule
    \end{tabularx}
    \caption{
        Comparison of selected best MAB algorithms from \Cref{tab:mab_comparison} using paired t-test on JSSP set.
    }
    \label{tab:map_mab_paired_JSSP}
\end{table}

\begin{table}[ht]
    \centering
    \begin{tabularx}{\linewidth}{l|*6{X}}
        \toprule
         & \textbf{\tgL} & \textbf{\tgLR} & \textbf{\egL} & \textbf{\egHR} & \textbf{\bgL} & \textbf{\bgHR} \\
        \midrule
        \textbf{\tgL} & -- &  &  &  &  &  \\
        \textbf{\tgLR} & \cellcolor{white} \textbf{-20.32$^{*}$} & -- &  &  &  &  \\
        \textbf{\egL} & \cellcolor{white} -17.17 & \cellcolor{white} \textbf{3.15$^{*}$} & -- &  &  &  \\
        \textbf{\egHR} & \cellcolor{white} \textbf{-24.26$^{*}$} & \cellcolor{white} \textbf{-3.94$^{*}$} & \cellcolor{white} \textbf{-7.08$^{*}$} & -- &  &  \\
        \textbf{\bgL} & \cellcolor{white} -18.55 & \cellcolor{white} \textbf{1.77$^{*}$} & \cellcolor{white} -1.38 & \cellcolor{white} \textbf{5.70$^{*}$} & -- &  \\
        \textbf{\bgHR} & \cellcolor{white} \textbf{-26.22$^{*}$} & \cellcolor{white} \textbf{-5.90$^{*}$} & \cellcolor{white} \textbf{-9.04$^{*}$} & \cellcolor{white} \textbf{-1.96$^{*}$} & \cellcolor{white} \textbf{-7.66$^{*}$} & -- \\
        \bottomrule
    \end{tabularx}
    \caption{
        Comparison of selected best MAB algorithms from \Cref{tab:mab_comparison} using paired t-test on RCPSP set.
    }
    \label{tab:map_mab_paired_RCPSP}
\end{table}

\begin{table}[ht]
    \centering
    \begin{tabularx}{\linewidth}{l|*6{X}}
        \toprule
         & \textbf{\tgL} & \textbf{\tgLR} & \textbf{\egL} & \textbf{\egHR} & \textbf{\bgL} & \textbf{\bgHR} \\
        \midrule
        \textbf{\tgL} & -- &  &  &  &  &  \\
        \textbf{\tgLR} & \cellcolor{white} \textbf{-24.41$^{*}$} & -- &  &  &  &  \\
        \textbf{\egL} & \cellcolor{white} \textbf{-8.81} & \cellcolor{white} \textbf{15.61$^{*}$} & -- &  &  &  \\
        \textbf{\egHR} & \cellcolor{white} \textbf{-23.14$^{*}$} & \cellcolor{white} \textbf{1.27} & \cellcolor{white} \textbf{-14.34$^{*}$} & -- &  &  \\
        \textbf{\bgL} & \cellcolor{white} -14.67 & \cellcolor{white} \textbf{9.74$^{*}$} & \cellcolor{white} \textbf{-5.87$^{*}$} & \cellcolor{white} \textbf{8.47$^{*}$} & -- &  \\
        \textbf{\bgHR} & \cellcolor{white} \textbf{-26.36$^{*}$} & \cellcolor{white} \textbf{-1.95$^{*}$} & \cellcolor{white} \textbf{-17.56$^{*}$} & \cellcolor{white} -3.22 & \cellcolor{white} \textbf{-11.69$^{*}$} & -- \\
        \bottomrule
    \end{tabularx}
    \caption{
        Comparison of selected best MAB algorithms from \Cref{tab:mab_comparison} using Wilcoxon signed rank test on JSSP set.
    }
    \label{tab:map_mab_wilcoxon_JSSP}
\end{table}

\begin{table}[ht]
    \centering
    \begin{tabularx}{\linewidth}{l|*6{X}}
        \toprule
         & \textbf{\tgL} & \textbf{\tgLR} & \textbf{\egL} & \textbf{\egHR} & \textbf{\bgL} & \textbf{\bgHR} \\
        \midrule
        \textbf{\tgL} & -- &  &  &  &  &  \\
        \textbf{\tgLR} & \cellcolor{white} \textbf{-20.32$^{*}$} & -- &  &  &  &  \\
        \textbf{\egL} & \cellcolor{white} -17.17 & \cellcolor{white} \textbf{3.15$^{*}$} & -- &  &  &  \\
        \textbf{\egHR} & \cellcolor{white} \textbf{-24.26$^{*}$} & \cellcolor{white} \textbf{-3.94$^{*}$} & \cellcolor{white} \textbf{-7.08$^{*}$} & -- &  &  \\
        \textbf{\bgL} & \cellcolor{white} -18.55 & \cellcolor{white} \textbf{1.77$^{*}$} & \cellcolor{white} -1.38 & \cellcolor{white} \textbf{5.70$^{*}$} & -- &  \\
        \textbf{\bgHR} & \cellcolor{white} \textbf{-26.22$^{*}$} & \cellcolor{white} \textbf{-5.90$^{*}$} & \cellcolor{white} \textbf{-9.04$^{*}$} & \cellcolor{white} \textbf{-1.96$^{*}$} & \cellcolor{white} \textbf{-7.66$^{*}$} & -- \\
        \bottomrule
    \end{tabularx}
    \caption{
        Comparison of selected best MAB algorithms from \Cref{tab:mab_comparison} using Wilcoxon signed rank test on RCPSP set.
    }
    \label{tab:map_mab_wilcoxon_RCPSP}
\end{table}

\clearpage

\subsection{All comparison}
The results in Tables \ref{tab:map_all_paired_JSSP}, \ref{tab:map_all_paired_RCPSP}, \ref{tab:map_all_wilcoxon_JSSP}, and \ref{tab:map_all_wilcoxon_RCPSP} demonstrate that the final configuration of row 5 in Table \ref{tab:all_comparison} is better than any other compared configuration with strong statistical significance (p $<$ 0.005) based on both the Wilcoxon and paired t test on both the JSSP and RCPSP datasets.

\begin{table}[ht]
    \centering
    \begin{tabularx}{\linewidth}{l|*6{X}}
        \toprule
         & \textbf{Row 1} & \textbf{Row 2} & \textbf{Row 3} & \textbf{Row 4} & \textbf{Row 5} & \textbf{Row 6} \\
        \midrule
        \textbf{Row 1} & -- &  &  &  &  &  \\
        \textbf{Row 2} & \cellcolor{white} -3.43 & -- &  &  &  &  \\
        \textbf{Row 3} & \cellcolor{white} -14.74 & \cellcolor{white} -11.31 & -- &  &  &  \\
        \textbf{Row 4} & \cellcolor{white} \textbf{-31.40$^{*}$} & \cellcolor{white} \textbf{-27.97$^{*}$} & \cellcolor{white} \textbf{-16.66$^{*}$} & -- &  &  \\
        \textbf{Row 5} & \cellcolor{white} \textbf{-38.03$^{*}$} & \cellcolor{white} \textbf{-34.60$^{*}$} & \cellcolor{white} \textbf{-23.29$^{*}$} & \cellcolor{white} \textbf{-6.63$^{*}$} & -- &  \\
        \textbf{Row 6} & \cellcolor{white} \textbf{103.73$^{*}$} & \cellcolor{white} \textbf{107.16$^{*}$} & \cellcolor{white} \textbf{118.48$^{*}$} & \cellcolor{white} \textbf{135.13$^{*}$} & \cellcolor{white} \textbf{141.76$^{*}$} & -- \\
        \bottomrule
    \end{tabularx}
    \caption{
        Comparison of selected configurations using paired t-test on JSSP set.
    }
    \label{tab:map_all_paired_JSSP}
\end{table}

\begin{table}[ht]
    \centering
    \begin{tabularx}{\linewidth}{l|*6{X}}
        \toprule
         & \textbf{Row 1} & \textbf{Row 2} & \textbf{Row 3} & \textbf{Row 4} & \textbf{Row 5} & \textbf{Row 6} \\
        \midrule
        \textbf{Row 1} & -- &  &  &  &  &  \\
        \textbf{Row 2} & \cellcolor{white} \textbf{-10.15} & -- &  &  &  &  \\
        \textbf{Row 3} & \cellcolor{white} \textbf{-8.14$^{*}$} & \cellcolor{white} 2.01 & -- &  &  &  \\
        \textbf{Row 4} & \cellcolor{white} \textbf{-17.20$^{*}$} & \cellcolor{white} \textbf{-7.06$^{*}$} & \cellcolor{white} \textbf{-9.07$^{*}$} & -- &  &  \\
        \textbf{Row 5} & \cellcolor{white} \textbf{-19.16$^{*}$} & \cellcolor{white} \textbf{-9.01$^{*}$} & \cellcolor{white} \textbf{-11.02$^{*}$} & \cellcolor{white} \textbf{-1.95$^{*}$} & -- &  \\
        \textbf{Row 6} & \cellcolor{white} \textbf{-4.75$^{*}$} & \cellcolor{white} \textbf{5.40$^{*}$} & \cellcolor{white} \textbf{3.38$^{*}$} & \cellcolor{white} \textbf{12.45$^{*}$} & \cellcolor{white} \textbf{14.40$^{*}$} & -- \\
        \bottomrule
    \end{tabularx}
    \caption{
        Comparison of selected configurations using paired t-test on RCPSP set.
    }
    \label{tab:map_all_paired_RCPSP}
\end{table}

\begin{table}[ht]
    \centering
    \begin{tabularx}{\linewidth}{l|*6{X}}
        \toprule
         & \textbf{Row 1} & \textbf{Row 2} & \textbf{Row 3} & \textbf{Row 4} & \textbf{Row 5} & \textbf{Row 6} \\
        \midrule
        \textbf{Row 1} & -- &  &  &  &  &  \\
        \textbf{Row 2} & \cellcolor{white} -3.43 & -- &  &  &  &  \\
        \textbf{Row 3} & \cellcolor{white} -14.74 & \cellcolor{white} -11.31 & -- &  &  &  \\
        \textbf{Row 4} & \cellcolor{white} \textbf{-31.40$^{*}$} & \cellcolor{white} \textbf{-27.97$^{*}$} & \cellcolor{white} \textbf{-16.66$^{*}$} & -- &  &  \\
        \textbf{Row 5} & \cellcolor{white} \textbf{-38.03$^{*}$} & \cellcolor{white} \textbf{-34.60$^{*}$} & \cellcolor{white} \textbf{-23.29$^{*}$} & \cellcolor{white} \textbf{-6.63$^{*}$} & -- &  \\
        \textbf{Row 6} & \cellcolor{white} \textbf{103.73$^{*}$} & \cellcolor{white} \textbf{107.16$^{*}$} & \cellcolor{white} \textbf{118.48$^{*}$} & \cellcolor{white} \textbf{135.13$^{*}$} & \cellcolor{white} \textbf{141.76$^{*}$} & -- \\
        \bottomrule
        \end{tabularx}
    \caption{
        Comparison of selected configurations using Wilcoxon signed rank test on JSSP set.
    }
    \label{tab:map_all_wilcoxon_JSSP}
\end{table}

\begin{table}[ht]
    \centering
    \begin{tabularx}{\linewidth}{l|*6{X}}
        \toprule
         & \textbf{Row 1} & \textbf{Row 2} & \textbf{Row 3} & \textbf{Row 4} & \textbf{Row 5} & \textbf{Row 6} \\
        \midrule
        \textbf{Row 1} & -- &  &  &  &  &  \\
        \textbf{Row 2} & \cellcolor{white} \textbf{-10.15} & -- &  &  &  &  \\
        \textbf{Row 3} & \cellcolor{white} \textbf{-8.14$^{*}$} & \cellcolor{white} 2.01 & -- &  &  &  \\
        \textbf{Row 4} & \cellcolor{white} \textbf{-17.20$^{*}$} & \cellcolor{white} \textbf{-7.06$^{*}$} & \cellcolor{white} \textbf{-9.07$^{*}$} & -- &  &  \\
        \textbf{Row 5} & \cellcolor{white} \textbf{-19.16$^{*}$} & \cellcolor{white} \textbf{-9.01$^{*}$} & \cellcolor{white} \textbf{-11.02$^{*}$} & \cellcolor{white} \textbf{-1.95$^{*}$} & -- &  \\
        \textbf{Row 6} & \cellcolor{white} \textbf{-4.75$^{*}$} & \cellcolor{white} \textbf{5.40$^{*}$} & \cellcolor{white} \textbf{3.38$^{*}$} & \cellcolor{white} \textbf{12.45$^{*}$} & \cellcolor{white} \textbf{14.40$^{*}$} & -- \\
        \bottomrule
    \end{tabularx}
    \caption{
        Comparison of selected configurations using Wilcoxon signed rank test on RCPSP set.
    }
    \label{tab:map_all_wilcoxon_RCPSP}
\end{table}

\FloatBarrier

\section{Results: Improved Bounds}
\label{app:list_bounds}
\newpage

\begin{singlespace}
\begin{tiny}
\begin{longtable}{lrrrr}
    \toprule
    \textbf{Name} & \textbf{Old LB \cite{optimizizer_jobshop,jjvh_jobshop} [-]}  & \textbf{New LB [-]}  & \textbf{Time [s]} & \textbf{Closed} \\
    \midrule
    \endfirsthead
    
    \multicolumn{5}{c}%
    {{\bfseries \tablename\ \thetable{} -- continued from previous page}} \\
    \toprule
    \textbf{Name} & \textbf{Old LB \cite{optimizizer_jobshop,jjvh_jobshop} [-]}  & \textbf{New LB [-]}  & \textbf{Time [s]} & \textbf{Closed} \\
    \midrule
    \endhead
    
    \midrule
    \multicolumn{5}{c}{{Continued on next page}} \\
    \endfoot
    
    \midrule
    \caption{List of improved lower bounds (New LB) compared to the existing state-of-the-art ones (Old LB) for open JSSP instances. Column "Closed" denotes, if the instance was closed, meaning the "New LB" equals to the optimal makespan.}
    \endlastfoot
        ta18	&1377	&1382	&819.032	&No\\
        ta22	&1561	&1576	&774.015	&No\\
        ta23	&1518	&1533	&713.053	&No\\
        ta25	&1558	&1575	&661.092	&No\\
        ta26	&1591	&1609	&856.031	&No\\
        ta27	&1652	&1665	&841.077	&No\\
        ta29	&1583	&1597	&746.097	&No\\
        ta30	&1528	&1539	&857.069	&No\\
        ta33	&1788	&1790	&382.069	&No\\
        ta40	&1651	&1652	&284.024	&No\\
        ta41	&1906	&1912	&733.042	&No\\
        ta42	&1884	&1887	&165.068	&No\\
        ta44	&1948	&1952	&603.011	&No\\
        ta46	&1957	&1966	&586.038	&No\\
        ta47	&1807	&1816	&594.049	&No\\
        ta48	&1912	&1915	&558.077	&No\\
        ta49	&1931	&1934	&618.024	&No\\
        ta50	&1833	&1837	&737.057	&No\\
        rcmax\_20\_15\_4	&2501	&2560	&755.083	&No\\
        rcmax\_20\_15\_10	&2651	&2706	&305.042	&Yes\\
        rcmax\_20\_15\_8	&2601	&2638	&783.065	&No\\
        rcmax\_20\_20\_6	&3042	&3188	&719.029	&No\\
        rcmax\_20\_20\_4	&2828	&2995	&733.091	&No\\
        rcmax\_20\_20\_7	&3051	&3188	&436.052	&Yes\\
        rcmax\_20\_20\_8	&2956	&3092	&242.054	&Yes\\
        rcmax\_20\_20\_5	&2858	&2984	&652.053	&Yes\\
        rcmax\_30\_15\_9	&3395	&3396	&246.097	&No\\
        rcmax\_30\_20\_10	&3709	&3717	&829.014	&No\\
        rcmax\_30\_20\_8	&3672	&3697	&686.017	&No\\
        rcmax\_30\_20\_2	&3604	&3619	&880.003	&No\\
        cscmax\_20\_15\_10	&3007	&3130	&396.026	&No\\
        cscmax\_20\_15\_5	&3224	&3321	&881.096	&No\\
        cscmax\_20\_15\_8	&3292	&3390	&797.027	&No\\
        cscmax\_20\_15\_7	&3299	&3386	&541.084	&No\\
        cscmax\_20\_15\_1	&3039	&3181	&219.043	&No\\
        cscmax\_20\_20\_6	&3575	&3740	&726.054	&No\\
        cscmax\_20\_20\_4	&3522	&3684	&723.058	&No\\
        cscmax\_20\_20\_3	&3447	&3583	&217.055	&No\\
        cscmax\_20\_20\_2	&3403	&3515	&861.068	&No\\
        cscmax\_20\_20\_9	&3496	&3588	&682.01	&No\\
        cscmax\_30\_15\_2	&3954	&4042	&61.008	&No\\
        cscmax\_30\_15\_9	&4094	&4170	&66.096	&No\\
        cscmax\_30\_15\_10	&4141	&4234	&682.007	&No\\
        cscmax\_30\_15\_5	&4202	&4271	&626.017	&No\\
        cscmax\_30\_15\_6	&4146	&4180	&857.076	&No\\
        cscmax\_30\_20\_9	&4554	&4729	&852.072	&No\\
        cscmax\_30\_20\_7	&4302	&4444	&797.021	&No\\
        cscmax\_30\_20\_3	&4319	&4449	&336.065	&No\\
        cscmax\_30\_20\_6	&4219	&4346	&848.088	&No\\
        cscmax\_30\_20\_4	&4319	&4447	&589.062	&No\\
        cscmax\_40\_15\_3	&4917	&5014	&289.033	&No\\
        cscmax\_40\_15\_6	&5041	&5160	&364.041	&No\\
        cscmax\_40\_15\_8	&5111	&5232	&813.029	&No\\
        cscmax\_40\_15\_4	&5130	&5148	&723.016	&No\\
        cscmax\_40\_15\_7	&5107	&5121	&2.005	&No\\
        cscmax\_40\_20\_10	&5397	&5512	&765.081	&No\\
        cscmax\_40\_20\_6	&5589	&5650	&237.068	&No\\
        cscmax\_40\_20\_8	&5426	&5493	&820.075	&No\\
        cscmax\_40\_20\_5	&5423	&5500	&549.063	&No\\
        cscmax\_40\_20\_9	&5501	&5619	&790.041	&No\\
        cscmax\_50\_15\_8	&6080	&6103	&5.056	&No\\
        cscmax\_50\_15\_6	&6395	&6430	&46.084	&No\\
        cscmax\_50\_15\_10	&6001	&6013	&0.092	&No\\
        cscmax\_50\_15\_4	&6123	&6128	&12.027	&No\\
        cscmax\_50\_15\_3	&6029	&6046	&337.027	&No\\
        cscmax\_50\_20\_1	&6342	&6463	&721.038	&No\\
        cscmax\_50\_20\_4	&6499	&6517	&103.042	&No\\
        cscmax\_50\_20\_3	&6586	&6638	&799.09	&No\\
        cscmax\_50\_20\_7	&6650	&6709	&797.005	&No\\
        abz8	&648	&652	&805.052	&No\\
        swv06	&1630	&1635	&843.091	&No\\
        swv07	&1513	&1523	&604.033	&No\\
        swv08	&1671	&1679	&847.005	&No\\
        swv09	&1633	&1638	&807.042	&No\\
        swv10	&1663	&1666	&164.01	&No\\
        yn2	&870	&879	&785.071	&No\\
        yn3	&859	&866	&690.082	&No\\
        yn4	&929	&935	&852.086	&No\\   
\end{longtable}
\end{tiny}
\label{tab:sota_instances_jssp}

\newpage
\begin{tiny}
\begin{longtable}{lrrrr}
    \toprule
    \textbf{Name} & \textbf{Old LB \cite{ugent_rcpsp} [-]}  & \textbf{New LB [-]}  & \textbf{Time [s]} & \textbf{Closed} \\
    \midrule
    \endfirsthead
    
    \multicolumn{5}{c}%
    {{\bfseries \tablename\ \thetable{} -- continued from previous page}} \\
    \toprule
    \textbf{Name} & \textbf{Old LB \cite{ugent_rcpsp} [-]}  & \textbf{New LB [-]}  & \textbf{Time [s]} & \textbf{Closed} \\
    \midrule
    \endhead
    
    \midrule
    \multicolumn{5}{c}{{Continued on next page}} \\
    \endfoot
    
    \midrule
    \caption{List of improved lower bounds (New LB) compared to the existing state-of-the-art ones (Old LB) for open RCPSP instances. Column "Closed" denotes, if the instance was closed, meaning the "New LB" equals to the optimal makespan.}
    \endlastfoot
    
        j1201\_1.sm	&104	&105	&11.052	&Yes\\
        j1206\_1.sm	&133	&137	&733.013	&No\\
        j1206\_2.sm	&125	&127	&19.093	&No\\
        j1206\_5.sm	&116	&118	&14.011	&No\\
        j1206\_6.sm	&140	&143	&130.097	&No\\
        j1206\_7.sm	&154	&155	&46.011	&No\\
        j1206\_8.sm	&140	&142	&21.074	&No\\
        j1206\_9.sm	&148	&151	&151.021	&No\\
        j1206\_10.sm	&157	&158	&10.098	&No\\
        j1207\_1.sm	&97	&99	&133.093	&No\\
        j1207\_4.sm	&106	&107	&38.031	&No\\
        j1207\_5.sm	&125	&127	&868.095	&No\\
        j1207\_6.sm	&115	&117	&155.02	&No\\
        j1207\_7.sm	&113	&115	&806.032	&No\\
        j1207\_8.sm	&92	&93	&1.04	&No\\
        j1207\_9.sm	&85	&87	&76.05	&No\\
        j1207\_10.sm	&111	&112	&28.027	&No\\
        j1208\_5.sm	&99	&101	&175.002	&No\\
        j1208\_9.sm	&89	&91	&253.046	&No\\
        j12011\_1.sm	&157	&159	&60.061	&No\\
        j12011\_2.sm	&147	&148	&835.056	&No\\
        j12011\_3.sm	&188	&190	&177.058	&No\\
        j12011\_4.sm	&177	&183	&224.06	&No\\
        j12011\_5.sm	&193	&195	&144.088	&No\\
        j12011\_6.sm	&192	&193	&436.037	&No\\
        j12011\_7.sm	&148	&152	&33.049	&No\\
        j12011\_8.sm	&152	&154	&149.058	&No\\
        j12011\_10.sm	&164	&166	&152.029	&No\\
        j12012\_1.sm	&125	&128	&406.064	&No\\
        j12012\_2.sm	&110	&112	&8.004	&No\\
        j12012\_3.sm	&132	&133	&869.061	&No\\
        j12012\_4.sm	&121	&122	&24.06	&No\\
        j12012\_5.sm	&154	&155	&2.064	&No\\
        j12012\_6.sm	&115	&116	&2.066	&No\\
        j12012\_8.sm	&113	&114	&470.085	&No\\
        j12012\_9.sm	&101	&102	&195.078	&No\\
        j12013\_1.sm	&123	&124	&220.097	&No\\
        j12013\_3.sm	&114	&116	&111.088	&No\\
        j12013\_4.sm	&108	&109	&101.051	&No\\
        j12013\_6.sm	&95	&96	&88.036	&No\\
        j12013\_9.sm	&82	&84	&312.027	&No\\
        j12014\_2.sm	&90	&91	&113.011	&No\\
        j12014\_7.sm	&89	&90	&2.034	&Yes\\
        j12016\_4.sm	&189	&191	&25.048	&No\\
        j12016\_5.sm	&184	&186	&89.007	&No\\
        j12016\_6.sm	&194	&195	&221.09	&No\\
        j12016\_10.sm	&203	&204	&12.034	&No\\
        j12017\_4.sm	&117	&118	&50.085	&No\\
        j12017\_9.sm	&129	&130	&203.059	&No\\
        j12018\_7.sm	&112	&113	&458.059	&No\\
        j12018\_8.sm	&101	&102	&1.01	&No\\
        j12018\_9.sm	&88	&89	&1.006	&No\\
        j12018\_10.sm	&96	&97	&5.077	&No\\
        j12026\_3.sm	&155	&161	&870.078	&No\\
        j12026\_5.sm	&139	&140	&38.005	&No\\
        j12026\_6.sm	&170	&180	&514.053	&No\\
        j12026\_7.sm	&147	&149	&494.056	&No\\
        j12026\_9.sm	&160	&162	&89.044	&No\\
        j12027\_1.sm	&105	&107	&1.084	&No\\
        j12027\_2.sm	&109	&111	&140.079	&No\\
        j12027\_3.sm	&141	&143	&705.082	&No\\
        j12027\_4.sm	&104	&105	&3.018	&No\\
        j12027\_5.sm	&105	&106	&487.054	&No\\
        j12027\_6.sm	&132	&136	&36.088	&No\\
        j12027\_7.sm	&118	&123	&863.095	&No\\
        j12027\_9.sm	&120	&122	&28.013	&No\\
        j12028\_1.sm	&105	&107	&744.023	&No\\
        j12031\_1.sm	&180	&183	&289.024	&No\\
        j12031\_2.sm	&175	&180	&287.016	&No\\
        j12031\_3.sm	&159	&161	&666.019	&No\\
        j12031\_4.sm	&194	&196	&330.058	&No\\
        j12031\_5.sm	&186	&187	&104.034	&No\\
        j12031\_6.sm	&181	&184	&403.071	&No\\
        j12031\_7.sm	&190	&195	&161.028	&No\\
        j12031\_8.sm	&176	&177	&197.068	&No\\
        j12031\_9.sm	&174	&177	&265.08	&No\\
        j12031\_10.sm	&201	&206	&625.087	&No\\
        j12032\_2.sm	&122	&123	&38.047	&No\\
        j12032\_3.sm	&134	&135	&733.071	&No\\
        j12032\_4.sm	&127	&128	&287.022	&No\\
        j12032\_5.sm	&131	&133	&1.069	&No\\
        j12032\_6.sm	&121	&123	&714.093	&No\\
        j12032\_7.sm	&118	&119	&278.073	&No\\
        j12032\_8.sm	&131	&132	&2.016	&No\\
        j12033\_1.sm	&104	&105	&3.094	&No\\
        j12033\_2.sm	&106	&107	&24.047	&No\\
        j12033\_3.sm	&101	&102	&3.01	&No\\
        j12033\_4.sm	&106	&107	&250.043	&No\\
        j12033\_9.sm	&109	&110	&427.026	&No\\
        j12033\_10.sm	&102	&103	&551.083	&No\\
        j12034\_2.sm	&102	&103	&3.014	&No\\
        j12034\_3.sm	&98	&100	&35.009	&No\\
        j12034\_5.sm	&101	&102	&272.009	&No\\
        j12036\_1.sm	&199	&201	&131.038	&No\\
        j12036\_9.sm	&202	&203	&259.077	&No\\
        j12036\_10.sm	&198	&199	&84.034	&No\\
        j12037\_2.sm	&140	&141	&11.06	&No\\
        j12037\_4.sm	&156	&157	&180.054	&No\\
        j12037\_5.sm	&194	&195	&65.086	&No\\
        j12037\_7.sm	&151	&152	&212.025	&No\\
        j12037\_8.sm	&168	&170	&336.086	&No\\
        j12037\_9.sm	&137	&138	&15.061	&No\\
        j12038\_1.sm	&105	&106	&275.053	&No\\
        j12038\_2.sm	&118	&119	&22.099	&No\\
        j12038\_3.sm	&153	&154	&192.072	&No\\
        j12038\_4.sm	&137	&138	&13.026	&No\\
        j12038\_6.sm	&118	&119	&77.018	&No\\
        j12038\_10.sm	&136	&137	&33.062	&No\\
        j12039\_2.sm	&104	&106	&225.092	&No\\
        j12039\_9.sm	&89	&90	&472.051	&No\\
        j12040\_1.sm	&79	&80	&20.099	&No\\
        j12046\_1.sm	&171	&175	&465.061	&No\\
        j12046\_4.sm	&160	&161	&635.015	&No\\
        j12046\_5.sm	&135	&141	&461.092	&No\\
        j12046\_7.sm	&157	&162	&256.09	&No\\
        j12046\_10.sm	&174	&184	&157.03	&No\\
        j12047\_3.sm	&118	&122	&47.086	&No\\
        j12047\_4.sm	&119	&123	&284.003	&No\\
        j12047\_6.sm	&127	&134	&582.073	&No\\
        j12047\_8.sm	&122	&127	&389.0	&No\\
        j12047\_10.sm	&126	&131	&470.097	&No\\
        j12048\_2.sm	&111	&112	&439.068	&Yes\\
        j12048\_4.sm	&120	&123	&5.049	&No\\
        j12048\_6.sm	&102	&103	&594.099	&No\\
        j12051\_2.sm	&200	&201	&133.086	&No\\
        j12051\_3.sm	&192	&198	&419.064	&No\\
        j12051\_4.sm	&196	&201	&158.047	&No\\
        j12051\_6.sm	&193	&198	&818.055	&No\\
        j12051\_7.sm	&185	&187	&143.05	&No\\
        j12051\_8.sm	&185	&187	&39.065	&No\\
        j12051\_9.sm	&191	&192	&346.044	&No\\
        j12051\_10.sm	&200	&203	&816.036	&No\\
        j12052\_1.sm	&160	&161	&582.035	&No\\
        j12052\_2.sm	&168	&172	&133.052	&No\\
        j12052\_3.sm	&125	&126	&70.069	&No\\
        j12052\_4.sm	&157	&158	&46.03	&No\\
        j12052\_5.sm	&158	&160	&833.027	&No\\
        j12052\_6.sm	&184	&187	&733.01	&No\\
        j12052\_7.sm	&140	&143	&15.047	&No\\
        j12052\_8.sm	&147	&149	&100.027	&No\\
        j12052\_9.sm	&141	&144	&364.072	&No\\
        j12052\_10.sm	&130	&135	&223.08	&No\\
        j12053\_1.sm	&136	&139	&89.057	&No\\
        j12053\_2.sm	&109	&110	&61.008	&No\\
        j12053\_3.sm	&105	&107	&529.07	&No\\
        j12053\_4.sm	&137	&138	&114.059	&No\\
        j12053\_5.sm	&109	&110	&213.076	&No\\
        j12053\_8.sm	&134	&135	&1.091	&No\\
        j12053\_10.sm	&123	&126	&37.004	&No\\
        j12054\_1.sm	&101	&102	&25.031	&No\\
        j12054\_5.sm	&106	&107	&37.075	&No\\
        j12054\_6.sm	&104	&105	&477.092	&No\\
        j12054\_9.sm	&104	&105	&66.091	&No\\
        j12055\_6.sm	&98	&99	&563.025	&No\\
        j12057\_1.sm	&172	&174	&651.059	&No\\
        j12057\_2.sm	&150	&152	&299.022	&No\\
        j12057\_5.sm	&169	&171	&290.027	&No\\
        j12057\_6.sm	&175	&177	&122.086	&No\\
        j12057\_7.sm	&155	&157	&722.073	&No\\
        j12057\_9.sm	&156	&159	&885.076	&No\\
        j12058\_1.sm	&133	&134	&109.058	&No\\
        j12058\_2.sm	&121	&122	&9.018	&No\\
        j12058\_3.sm	&116	&117	&43.09	&No\\
        j12058\_4.sm	&137	&139	&8.015	&No\\
        j12058\_5.sm	&115	&116	&6.037	&No\\
        j12058\_6.sm	&134	&136	&497.094	&No\\
        j12058\_7.sm	&142	&143	&10.029	&No\\
        j12058\_9.sm	&125	&127	&556.049	&No\\
        j12059\_2.sm	&103	&104	&271.025	&No\\
        j12059\_8.sm	&106	&107	&724.024	&No\\
        j12059\_9.sm	&116	&117	&1.029	&No\\
        j12059\_10.sm	&127	&128	&541.077	&No\\
        j12060\_3.sm	&87	&88	&2.018	&No\\
        j609\_1.sm	&84	&85	&35.0	&No\\
        j609\_6.sm	&105	&106	&877.093	&No\\
        j609\_7.sm	&103	&105	&745.009	&No\\
        j6013\_1.sm	&105	&106	&345.008	&No\\
        j6013\_9.sm	&95	&98	&836.042	&No\\
        j6013\_10.sm	&112	&114	&4.092	&No\\
        j6025\_6.sm	&105	&106	&861.058	&No\\
        j6029\_1.sm	&98	&99	&267.005	&No\\
        j6029\_8.sm	&96	&98	&38.093	&No\\
        j6041\_3.sm	&90	&91	&692.011	&No\\
        j6045\_6.sm	&133	&135	&432.039	&No\\
        j6045\_10.sm	&105	&106	&411.07	&No\\
        j905\_3.sm	&83	&85	&25.088	&No\\
        j905\_8.sm	&96	&99	&429.023	&No\\
        j905\_10.sm	&94	&95	&127.09	&No\\
        j909\_2.sm	&121	&122	&88.028	&No\\
        j909\_4.sm	&120	&121	&279.013	&No\\
        j909\_5.sm	&127	&131	&577.071	&No\\
        j909\_7.sm	&102	&103	&9.015	&No\\
        j909\_8.sm	&110	&111	&27.063	&No\\
        j9013\_1.sm	&129	&130	&564.051	&No\\
        j9013\_2.sm	&118	&121	&576.068	&No\\
        j9013\_3.sm	&104	&105	&48.065	&No\\
        j9013\_5.sm	&108	&109	&81.092	&No\\
        j9013\_8.sm	&112	&113	&4.073	&No\\
        j9013\_9.sm	&117	&118	&77.009	&No\\
        j9013\_10.sm	&112	&114	&294.074	&No\\
        j9025\_1.sm	&116	&117	&36.073	&No\\
        j9025\_2.sm	&123	&124	&356.039	&No\\
        j9025\_4.sm	&127	&130	&178.021	&No\\
        j9025\_5.sm	&109	&110	&4.05	&No\\
        j9025\_6.sm	&112	&115	&371.037	&No\\
        j9025\_9.sm	&97	&100	&110.064	&No\\
        j9025\_10.sm	&119	&121	&554.076	&No\\
        j9029\_1.sm	&125	&128	&232.06	&No\\
        j9029\_3.sm	&136	&137	&716.029	&No\\
        j9029\_6.sm	&116	&118	&344.099	&No\\
        j9029\_7.sm	&160	&161	&302.095	&No\\
        j9029\_8.sm	&145	&147	&487.078	&No\\
        j9029\_10.sm	&118	&119	&250.027	&No\\
        j9030\_9.sm	&91	&92	&85.02	&No\\
        j9041\_1.sm	&128	&133	&142.084	&No\\
        j9041\_5.sm	&117	&119	&778.054	&No\\
        j9041\_6.sm	&123	&129	&355.031	&No\\
        j9041\_8.sm	&147	&150	&473.062	&No\\
        j9041\_9.sm	&110	&111	&183.037	&No\\
        j9045\_1.sm	&142	&144	&35.007	&No\\
        j9045\_2.sm	&137	&139	&78.056	&No\\
        j9045\_4.sm	&125	&126	&143.091	&No\\
        j9045\_7.sm	&129	&130	&382.08	&No\\
        j9045\_8.sm	&148	&151	&57.034	&No\\
        j9045\_9.sm	&144	&146	&665.002	&No\\
        j9045\_10.sm	&155	&157	&435.061	&No\\
\end{longtable}
\end{tiny}
\label{tab:sota_instances_rcpsp}
\end{singlespace}


\newpage

\begin{singlespace}
\bibliographystyle{elsarticle-num} 
\bibliography{paper}

\begin{thebibliography}{10}
\expandafter\ifx\csname url\endcsname\relax
  \def\url#1{\texttt{#1}}\fi
\expandafter\ifx\csname urlprefix\endcsname\relax\def\urlprefix{URL }\fi
\expandafter\ifx\csname href\endcsname\relax
  \def\href#1#2{#2} \def\path#1{#1}\fi

\bibitem{LIESS}
O.~Liess, P.~Michelon, \href{https://link.springer.com/article/10.1007/s10479-007-0188-y}{A constraint programming approach for the resource-constrained project scheduling problem}, Annals of Operations Research 157 (2008).
\newblock \href {https://doi.org/https://doi.org/10.1007/s10479-007-0188-y} {\path{doi:https://doi.org/10.1007/s10479-007-0188-y}}.
\newline\urlprefix\url{https://link.springer.com/article/10.1007/s10479-007-0188-y}

\bibitem{HE20123331}
F.~He, R.~Qu, \href{https://www.sciencedirect.com/science/article/pii/S0305054812000986}{A constraint programming based column generation approach to nurse rostering problems}, Computers \& Operations Research 39~(12) (2012) 3331--3343.
\newblock \href {https://doi.org/https://doi.org/10.1016/j.cor.2012.04.018} {\path{doi:https://doi.org/10.1016/j.cor.2012.04.018}}.
\newline\urlprefix\url{https://www.sciencedirect.com/science/article/pii/S0305054812000986}

\bibitem{icores25}
L.~Nedbálek, A.~Novák, Bottleneck identification in resource-constrained project scheduling via constraint relaxation, in: Proceedings of the 14th International Conference on Operations Research and Enterprise Systems - ICORES, INSTICC, SciTePress, 2025, pp. 340--347.
\newblock \href {https://doi.org/10.5220/0013253700003893} {\path{doi:10.5220/0013253700003893}}.

\bibitem{ABREU2022108128}
L.~R. Abreu, M.~S. Nagano, \href{https://www.sciencedirect.com/science/article/pii/S036083522200198X}{A new hybridization of adaptive large neighborhood search with constraint programming for open shop scheduling with sequence-dependent setup times}, Computers \& Industrial Engineering 168 (2022) 108128.
\newblock \href {https://doi.org/https://doi.org/10.1016/j.cie.2022.108128} {\path{doi:https://doi.org/10.1016/j.cie.2022.108128}}.
\newline\urlprefix\url{https://www.sciencedirect.com/science/article/pii/S036083522200198X}

\bibitem{ROHANINEJAD2023108958}
M.~Rohaninejad, Z.~Hanzálek, \href{https://www.sciencedirect.com/science/article/pii/S0925527323001901}{Multi-level lot-sizing and job shop scheduling with lot-streaming: Reformulation and solution approaches}, International Journal of Production Economics 263 (2023) 108958.
\newblock \href {https://doi.org/https://doi.org/10.1016/j.ijpe.2023.108958} {\path{doi:https://doi.org/10.1016/j.ijpe.2023.108958}}.
\newline\urlprefix\url{https://www.sciencedirect.com/science/article/pii/S0925527323001901}

\bibitem{HEINZ2022108586}
V.~Heinz, A.~Novák, M.~Vlk, Z.~Hanzálek, \href{https://www.sciencedirect.com/science/article/pii/S0360835222005836}{Constraint programming and constructive heuristics for parallel machine scheduling with sequence-dependent setups and common servers}, Computers \& Industrial Engineering 172 (2022) 108586.
\newblock \href {https://doi.org/https://doi.org/10.1016/j.cie.2022.108586} {\path{doi:https://doi.org/10.1016/j.cie.2022.108586}}.
\newline\urlprefix\url{https://www.sciencedirect.com/science/article/pii/S0360835222005836}

\bibitem{LUNARDI2020105020}
W.~T. Lunardi, E.~G. Birgin, P.~Laborie, D.~P. Ronconi, H.~Voos, \href{https://www.sciencedirect.com/science/article/pii/S0305054820301374}{Mixed integer linear programming and constraint programming models for the online printing shop scheduling problem}, Computers \& Operations Research 123 (2020) 105020.
\newblock \href {https://doi.org/https://doi.org/10.1016/j.cor.2020.105020} {\path{doi:https://doi.org/10.1016/j.cor.2020.105020}}.
\newline\urlprefix\url{https://www.sciencedirect.com/science/article/pii/S0305054820301374}

\bibitem{FATEMIANARAKI2023102770}
S.~Fatemi-Anaraki, R.~Tavakkoli-Moghaddam, M.~Foumani, B.~Vahedi-Nouri, \href{https://www.sciencedirect.com/science/article/pii/S0305048322001773}{Scheduling of multi-robot job shop systems in dynamic environments: Mixed-integer linear programming and constraint programming approaches}, Omega 115 (2023) 102770.
\newblock \href {https://doi.org/https://doi.org/10.1016/j.omega.2022.102770} {\path{doi:https://doi.org/10.1016/j.omega.2022.102770}}.
\newline\urlprefix\url{https://www.sciencedirect.com/science/article/pii/S0305048322001773}

\bibitem{vilim2015failure}
P.~Vil{\'\i}m, P.~Laborie, P.~Shaw, Failure-directed search for constraint-based scheduling, in: International Conference on Integration of Constraint Programming, Artificial Intelligence, and Operations Research, Springer, 2015, pp. 437--453.

\bibitem{ibm_cp_optimizer}
{IBM}, {CP Optimizer}, \url{https://www.ibm.com/analytics/cplex-cp-optimizer}, accessed: April 17, 2023 (2023).

\bibitem{naderia2022mixed}
B.~Naderia, R.~Ruizb, V.~Roshanaeic, Mixed-integer programming versus constraint programming for shop scheduling problems: New results and outlook, INFORMS Journal on Computing (2023).

\bibitem{hauder2020resource}
V.~A. Hauder, A.~Beham, S.~Raggl, S.~N. Parragh, M.~Affenzeller, Resource-constrained multi-project scheduling with activity and time flexibility, Computers \& Industrial Engineering 150 (2020) 106857.

\bibitem{BRAILSFORD1999557}
S.~C. Brailsford, C.~N. Potts, B.~M. Smith, \href{https://www.sciencedirect.com/science/article/pii/S0377221798003646}{Constraint satisfaction problems: Algorithms and applications}, European Journal of Operational Research 119~(3) (1999) 557--581.
\newblock \href {https://doi.org/https://doi.org/10.1016/S0377-2217(98)00364-6} {\path{doi:https://doi.org/10.1016/S0377-2217(98)00364-6}}.
\newline\urlprefix\url{https://www.sciencedirect.com/science/article/pii/S0377221798003646}

\bibitem{bams/1183517370}
H.~Robbins, {Some aspects of the sequential design of experiments}, Bulletin of the American Mathematical Society 58~(5) (1952) 527 -- 535.

\bibitem{Kazikova_Pluhacek_Senkerik_2020}
A.~Kazikova, M.~Pluhacek, R.~Senkerik, \href{https://mendel-journal.org/index.php/mendel/article/view/120}{Why tuning the control parameters of metaheuristic algorithms is so important for fair comparison?}, MENDEL 26~(2) (2020) 9--16.
\newblock \href {https://doi.org/10.13164/mendel.2020.2.009} {\path{doi:10.13164/mendel.2020.2.009}}.
\newline\urlprefix\url{https://mendel-journal.org/index.php/mendel/article/view/120}

\bibitem{10.1007/978-3-540-70881-0_24}
C.~H.~A. Koster, J.~G. Beney, On the importance of parameter tuning in text categorization, in: I.~Virbitskaite, A.~Voronkov (Eds.), Perspectives of Systems Informatics, Springer Berlin Heidelberg, Berlin, Heidelberg, 2007, pp. 270--283.

\bibitem{refalo2004impact}
P.~Refalo, Impact-based search strategies for constraint programming, in: International Conference on Principles and Practice of Constraint Programming, Springer, 2004, pp. 557--571.

\bibitem{michel2012activity}
L.~Michel, P.~V. Hentenryck, Activity-based search for black-box constraint programming solvers, in: International Conference on Integration of Artificial Intelligence (AI) and Operations Research (OR) Techniques in Constraint Programming, Springer, 2012, pp. 228--243.

\bibitem{LECOUTRE20091592}
C.~Lecoutre, L.~Saïs, S.~Tabary, V.~Vidal, \href{https://www.sciencedirect.com/science/article/pii/S0004370209001040}{Reasoning from last conflict(s) in constraint programming}, Artificial Intelligence 173~(18) (2009) 1592--1614.
\newblock \href {https://doi.org/https://doi.org/10.1016/j.artint.2009.09.002} {\path{doi:https://doi.org/10.1016/j.artint.2009.09.002}}.
\newline\urlprefix\url{https://www.sciencedirect.com/science/article/pii/S0004370209001040}

\bibitem{boussemart2004boosting}
F.~Boussemart, F.~Hemery, C.~Lecoutre, L.~Sais, Boosting systematic search by weighting constraints, in: ECAI, Vol.~16, 2004, p. 146.

\bibitem{8995307}
H.~Wattez, C.~Lecoutre, A.~Paparrizou, S.~Tabary, Refining constraint weighting, in: 2019 IEEE 31st International Conference on Tools with Artificial Intelligence (ICTAI), 2019, pp. 71--77.
\newblock \href {https://doi.org/10.1109/ICTAI.2019.00019} {\path{doi:10.1109/ICTAI.2019.00019}}.

\bibitem{habet:hal-02090610}
D.~Habet, C.~Terrioux, \href{https://hal-amu.archives-ouvertes.fr/hal-02090610}{{Conflict History Based Branching Heuristic for CSP Solving}}, in: {Proceedings of the 8th International Workshop on Combinations of Intelligent Methods and Applications (CIMA)}, Volos, Greece, 2018, pp. 1--10.
\newline\urlprefix\url{https://hal-amu.archives-ouvertes.fr/hal-02090610}

\bibitem{Doolaard2022}
F.~Doolaard, N.~Yorke-Smith, \href{https://doi.org/10.1007/s10472-022-09816-z}{Online learning of variable ordering heuristics for constraint optimisation problems}, Annals of Mathematics and Artificial Intelligence (Oct 2022).
\newblock \href {https://doi.org/10.1007/s10472-022-09816-z} {\path{doi:10.1007/s10472-022-09816-z}}.
\newline\urlprefix\url{https://doi.org/10.1007/s10472-022-09816-z}

\bibitem{Xu_Wu_Li_Yin_2025}
J.~Xu, Y.~Wu, H.~Li, M.~Yin, \href{https://ojs.aaai.org/index.php/AAAI/article/view/33239}{Prediction-based adaptive variable ordering heuristics for constraint satisfaction problems}, Proceedings of the AAAI Conference on Artificial Intelligence 39~(11) (2025) 11390--11398.
\newblock \href {https://doi.org/10.1609/aaai.v39i11.33239} {\path{doi:10.1609/aaai.v39i11.33239}}.
\newline\urlprefix\url{https://ojs.aaai.org/index.php/AAAI/article/view/33239}

\bibitem{Zarpellon_Jo_Lodi_Bengio_2021}
G.~Zarpellon, J.~Jo, A.~Lodi, Y.~Bengio, \href{https://ojs.aaai.org/index.php/AAAI/article/view/16512}{Parameterizing branch-and-bound search trees to learn branching policies}, Proceedings of the AAAI Conference on Artificial Intelligence 35~(5) (2021) 3931--3939.
\newblock \href {https://doi.org/10.1609/aaai.v35i5.16512} {\path{doi:10.1609/aaai.v35i5.16512}}.
\newline\urlprefix\url{https://ojs.aaai.org/index.php/AAAI/article/view/16512}

\bibitem{Khalil_Le_Bodic_Song_Nemhauser_Dilkina_2016}
E.~Khalil, P.~Le~Bodic, L.~Song, G.~Nemhauser, B.~Dilkina, \href{https://ojs.aaai.org/index.php/AAAI/article/view/10080}{Learning to branch in mixed integer programming}, Proceedings of the AAAI Conference on Artificial Intelligence 30~(1) (Feb. 2016).
\newblock \href {https://doi.org/10.1609/aaai.v30i1.10080} {\path{doi:10.1609/aaai.v30i1.10080}}.
\newline\urlprefix\url{https://ojs.aaai.org/index.php/AAAI/article/view/10080}

\bibitem{BOUSKA2023990}
M.~Bouška, P.~Šůcha, A.~Novák, Z.~Hanzálek, \href{https://www.sciencedirect.com/science/article/pii/S0377221722008918}{Deep learning-driven scheduling algorithm for a single machine problem minimizing the total tardiness}, European Journal of Operational Research 308~(3) (2023) 990--1006.
\newblock \href {https://doi.org/https://doi.org/10.1016/j.ejor.2022.11.034} {\path{doi:https://doi.org/10.1016/j.ejor.2022.11.034}}.
\newline\urlprefix\url{https://www.sciencedirect.com/science/article/pii/S0377221722008918}

\bibitem{10.1007/978-3-319-18008-3_6}
A.~Bonfietti, M.~Lombardi, M.~Milano, Embedding decision trees and random forests in constraint programming, in: L.~Michel (Ed.), Integration of AI and OR Techniques in Constraint Programming, Springer International Publishing, Cham, 2015, pp. 74--90.

\bibitem{10.1007/978-3-030-78230-6_25}
F.~Chalumeau, I.~Coulon, Q.~Cappart, L.-M. Rousseau, Seapearl: A constraint programming solver guided by reinforcement learning, in: P.~J. Stuckey (Ed.), Integration of Constraint Programming, Artificial Intelligence, and Operations Research, Springer International Publishing, Cham, 2021, pp. 392--409.

\bibitem{cappart2020combining}
Q.~Cappart, T.~Moisan, L.-M. Rousseau, I.~Prémont-Schwarz, A.~Cire, Combining reinforcement learning and constraint programming for combinatorial optimization (2020).
\newblock \href {http://arxiv.org/abs/2006.01610} {\path{arXiv:2006.01610}}.

\bibitem{10.1007/978-3-319-40970-2_9}
J.~H. Liang, V.~Ganesh, P.~Poupart, K.~Czarnecki, Learning rate based branching heuristic for {SAT} solvers, in: N.~Creignou, D.~Le~Berre (Eds.), Theory and Applications of Satisfiability Testing -- {SAT} 2016, Springer International Publishing, Cham, 2016, pp. 123--140.

\bibitem{Popescu2022}
A.~Popescu, S.~Polat-Erdeniz, A.~Felfernig, M.~Uta, M.~Atas, V.-M. Le, K.~Pilsl, M.~Enzelsberger, T.~N.~T. Tran, \href{https://doi.org/10.1007/s10844-021-00666-5}{An overview of machine learning techniques in constraint solving}, Journal of Intelligent Information Systems 58~(1) (2022) 91--118.
\newblock \href {https://doi.org/10.1007/s10844-021-00666-5} {\path{doi:10.1007/s10844-021-00666-5}}.
\newline\urlprefix\url{https://doi.org/10.1007/s10844-021-00666-5}

\bibitem{Xia_Yap_2018}
W.~Xia, R.~Yap, \href{https://ojs.aaai.org/index.php/AAAI/article/view/12211}{Learning robust search strategies using a bandit-based approach}, Proceedings of the AAAI Conference on Artificial Intelligence 32~(1) (Apr. 2018).
\newblock \href {https://doi.org/10.1609/aaai.v32i1.12211} {\path{doi:10.1609/aaai.v32i1.12211}}.
\newline\urlprefix\url{https://ojs.aaai.org/index.php/AAAI/article/view/12211}

\bibitem{koriche2022best}
F.~Koriche, C.~Lecoutre, A.~Paparrizou, H.~Wattez, Best heuristic identification for constraint satisfaction, in: 31st International Joint Conference on Artificial Intelligence (IJCAI'22), 2022, pp. 1859--1865.

\bibitem{Kletzander_Musliu_2023}
L.~Kletzander, N.~Musliu, \href{https://ojs.aaai.org/index.php/AAAI/article/view/26466}{Large-state reinforcement learning for hyper-heuristics}, Proceedings of the AAAI Conference on Artificial Intelligence 37~(10) (2023) 12444--12452.
\newblock \href {https://doi.org/10.1609/aaai.v37i10.26466} {\path{doi:10.1609/aaai.v37i10.26466}}.
\newline\urlprefix\url{https://ojs.aaai.org/index.php/AAAI/article/view/26466}

\bibitem{Wattez2020Learning}
H.~Wattez, F.~Koriche, C.~Lecoutre, A.~Paparrizou, S.~Tabary, \href{https://hal.archives-ouvertes.fr/hal-03096124}{Learning variable ordering heuristics with multi-armed bandits and restarts}, in: G.~Gottlob, T.~Soininen, C.~Vieira, M.~Virkki (Eds.), ECAI 2020 - 24th European Conference on Artificial Intelligence, Vol. 325 of Frontiers in Artificial Intelligence and Applications, IOS Press, Santiago de Compostela (virtual), Spain, 2020, pp. 479--486, \url{https://doi.org/10.3233/FAIA200115}.
\newblock \href {https://doi.org/10.3233/FAIA200115} {\path{doi:10.3233/FAIA200115}}.
\newline\urlprefix\url{https://hal.archives-ouvertes.fr/hal-03096124}

\bibitem{balafrej2015mab}
A.~Balafrej, C.~Bessiere, A.~Paparrizou, \href{https://hal.archives-ouvertes.fr/hal-01234361}{Multi-armed bandits for adaptive constraint propagation}, in: Proceedings of the 24th International Joint Conference on Artificial Intelligence (IJCAI), AAAI Press, Buenos Aires, Argentina, 2015, pp. 290--296.
\newline\urlprefix\url{https://hal.archives-ouvertes.fr/hal-01234361}

\bibitem{loth2013bandit}
M.~Loth, M.~Sebag, Y.~Hamadi, M.~Schoenauer, Bandit-based search for constraint programming, in: International Conference on Principles and Practice of Constraint Programming, Springer, 2013, pp. 464--480.

\bibitem{Mahajan2008}
A.~Mahajan, D.~Teneketzis, \href{https://doi.org/10.1007/978-0-387-49819-5\_6}{Multi-Armed Bandit Problems}, Springer US, Boston, MA, 2008, Ch.~6, pp. 121--151.
\newblock \href {https://doi.org/10.1007/978-0-387-49819-5\_6} {\path{doi:10.1007/978-0-387-49819-5\_6}}.
\newline\urlprefix\url{https://doi.org/10.1007/978-0-387-49819-5\_6}

\bibitem{sutton2018reinforcement}
R.~S. Sutton, A.~G. Barto, Reinforcement learning: An introduction, MIT press, 2018.

\bibitem{DBLP:journals/corr/KuleshovP14}
V.~Kuleshov, D.~Precup, \href{http://arxiv.org/abs/1402.6028}{Algorithms for multi-armed bandit problems}, CoRR abs/1402.6028 (2014).
\newblock \href {http://arxiv.org/abs/1402.6028} {\path{arXiv:1402.6028}}.
\newline\urlprefix\url{http://arxiv.org/abs/1402.6028}

\bibitem{NIPS2011_e53a0a29}
O.~Chapelle, L.~Li, \href{https://proceedings.neurips.cc/paper\_files/paper/2011/file/e53a0a2978c28872a4505bdb51db06dc-Paper.pdf}{An empirical evaluation of thompson sampling}, in: J.~Shawe-Taylor, R.~Zemel, P.~Bartlett, F.~Pereira, K.~Weinberger (Eds.), Advances in Neural Information Processing Systems, Vol.~24, Curran Associates, Inc., 2011, pp. 2249--2257.
\newline\urlprefix\url{https://proceedings.neurips.cc/paper\_files/paper/2011/file/e53a0a2978c28872a4505bdb51db06dc-Paper.pdf}

\bibitem{optuna_2019}
T.~Akiba, S.~Sano, T.~Yanase, T.~Ohta, M.~Koyama, Optuna: A next-generation hyperparameter optimization framework, in: Proceedings of the 25th {ACM} {SIGKDD} International Conference on Knowledge Discovery and Data Mining, 2019, pp. 1--10.

\bibitem{TAILLARD1993278}
E.~Taillard, \href{https://www.sciencedirect.com/science/article/pii/037722179390182M}{Benchmarks for basic scheduling problems}, European Journal of Operational Research 64~(2) (1993) 278--285, project Management anf Scheduling.
\newblock \href {https://doi.org/https://doi.org/10.1016/0377-2217(93)90182-M} {\path{doi:https://doi.org/10.1016/0377-2217(93)90182-M}}.
\newline\urlprefix\url{https://www.sciencedirect.com/science/article/pii/037722179390182M}

\bibitem{DEMIRKOL1998137}
E.~Demirkol, S.~Mehta, R.~Uzsoy, \href{https://www.sciencedirect.com/science/article/pii/S0377221797000192}{Benchmarks for shop scheduling problems}, European Journal of Operational Research 109~(1) (1998) 137--141.
\newblock \href {https://doi.org/https://doi.org/10.1016/S0377-2217(97)00019-2} {\path{doi:https://doi.org/10.1016/S0377-2217(97)00019-2}}.
\newline\urlprefix\url{https://www.sciencedirect.com/science/article/pii/S0377221797000192}

\bibitem{10.2307/2632051}
J.~Adams, E.~Balas, D.~Zawack, \href{http://www.jstor.org/stable/2632051}{The shifting bottleneck procedure for job shop scheduling}, Management Science 34~(3) (1988) 391--401.
\newline\urlprefix\url{http://www.jstor.org/stable/2632051}

\bibitem{10.2307/2632676}
R.~H. Storer, S.~D. Wu, R.~Vaccari, \href{http://www.jstor.org/stable/2632676}{New search spaces for sequencing problems with application to job shop scheduling}, Management Science 38~(10) (1992) 1495--1509.
\newline\urlprefix\url{http://www.jstor.org/stable/2632676}

\bibitem{inproceedings}
T.~Yamada, R.~Nakano, A genetic algorithm applicable to large-scale job-shop problems., in: Parallel Problem Solving from Nature 2, Vol.~2, 1992, pp. 283--292.

\bibitem{KOLISCH1997205}
R.~Kolisch, A.~Sprecher, \href{https://www.sciencedirect.com/science/article/pii/S0377221796001701}{Psplib - a project scheduling problem library: Or software - orsep operations research software exchange program}, European Journal of Operational Research 96~(1) (1997) 205--216.
\newblock \href {https://doi.org/https://doi.org/10.1016/S0377-2217(96)00170-1} {\path{doi:https://doi.org/10.1016/S0377-2217(96)00170-1}}.
\newline\urlprefix\url{https://www.sciencedirect.com/science/article/pii/S0377221796001701}

\bibitem{optalcp}
ScheduleOpt, \href{https://scheduleopt.com/}{Optalcp}, Online, optalCP's solver landing page (2023).
\newline\urlprefix\url{https://scheduleopt.com/}

\bibitem{fds_experimental_results}
Optal, \href{https://gitlab.com/optal\_solver/fds\_results}{Fds experimental results}, Online, git repository (2023).
\newline\urlprefix\url{https://gitlab.com/optal\_solver/fds\_results}

\bibitem{optimizizer_jobshop}
Optimizizer, \href{https://optimizizer.com/jobshop.php}{Job shop scheduling problem solver}, Online, accessed: April 14, 2023 (n.d.).
\newline\urlprefix\url{https://optimizizer.com/jobshop.php}

\bibitem{jjvh_jobshop}
{Jongejan, J.V.}, \href{http://jobshop.jjvh.nl/}{Job shop problem solver}, Online, accessed: April 14, 2023 (n.d.).
\newline\urlprefix\url{http://jobshop.jjvh.nl/}

\bibitem{ugent_rcpsp}
U.~of~Ghent, \href{https://www.projectmanagement.ugent.be/research/project\_scheduling/rcpsp/}{Resource-constrained project scheduling}, Online, accessed: Mar 2, 2023 (n.d.).
\newline\urlprefix\url{https://www.projectmanagement.ugent.be/research/project\_scheduling/rcpsp/}

\bibitem{brinkkotter2001}
W.~Brinkkötter, P.~Brucker, Solving open benchmark instances for the job-shop problem by parallel head–tail adjustments, Journal of Scheduling 4~(1) (2001) 53--64, \url{https://onlinelibrary.wiley.com/doi/abs/10.1002/1099-1425(200101/02)4:1<53::AID-JOS59>3.0.CO;2-Y}.
\newblock \href {https://doi.org/10.1002/1099-1425(200101/02)4:1<53::AID-JOS59>3.0.CO;2-Y} {\path{doi:10.1002/1099-1425(200101/02)4:1<53::AID-JOS59>3.0.CO;2-Y}}.

\bibitem{YURASZECK2025106964}
F.~Yuraszeck, G.~Mejía, D.~A. Rossit, A.~Lüer-Villagra, \href{https://www.sciencedirect.com/science/article/pii/S0305054824004362}{A constraint programming-based lower bounding procedure for the job shop scheduling problem}, Computers \& Operations Research 177 (2025) 106964.
\newblock \href {https://doi.org/https://doi.org/10.1016/j.cor.2024.106964} {\path{doi:https://doi.org/10.1016/j.cor.2024.106964}}.
\newline\urlprefix\url{https://www.sciencedirect.com/science/article/pii/S0305054824004362}

\bibitem{COELHO2020104976}
J.~Coelho, M.~Vanhoucke, \href{https://www.sciencedirect.com/science/article/pii/S0305054820300939}{Going to the core of hard resource-constrained project scheduling instances}, Computers \& Operations Research 121 (2020) 104976.
\newblock \href {https://doi.org/https://doi.org/10.1016/j.cor.2020.104976} {\path{doi:https://doi.org/10.1016/j.cor.2020.104976}}.
\newline\urlprefix\url{https://www.sciencedirect.com/science/article/pii/S0305054820300939}

\bibitem{shaw1998using}
P.~Shaw, Using constraint programming and local search methods to solve vehicle routing problems, in: International conference on principles and practice of constraint programming, Springer, 1998, pp. 417--431.

\bibitem{godard2005randomized}
D.~Godard, P.~Laborie, W.~Nuijten, Randomized large neighborhood search for cumulative scheduling., in: ICAPS, Vol.~5, 2005, pp. 81--89.

\bibitem{edge_finding_vilim}
P.~Vilím, Edge finding filtering algorithm for discrete cumulative resources in ${O}(kn \log n)$, in: Principles and Practice of Constraint Programming - CP 2009, 2009, pp. 802--816.
\newblock \href {https://doi.org/10.1007/978-3-642-04244-7_62} {\path{doi:10.1007/978-3-642-04244-7_62}}.

\bibitem{10.1007/978-3-319-23219-5_11}
S.~Gay, R.~Hartert, P.~Schaus, Simple and scalable time-table filtering for the cumulative constraint, in: G.~Pesant (Ed.), Principles and Practice of Constraint Programming, Springer International Publishing, Cham, 2015, pp. 149--157.

\bibitem{10.1007/978-3-642-40627-0_42}
P.~Ouellet, C.-G. Quimper, Time-table extended-edge-finding for the cumulative constraint, in: C.~Schulte (Ed.), Principles and Practice of Constraint Programming, Springer Berlin Heidelberg, Berlin, Heidelberg, 2013, pp. 562--577.

\bibitem{vilim_disertace}
R.~P. Vilím, \href{https://vilim.eu/petr/disertace.pdf}{Global constraints in scheduling}, Online, accessed: October 16, 2023 (n.d.).
\newline\urlprefix\url{https://vilim.eu/petr/disertace.pdf}

\bibitem{baptiste2001}
W.~N. Philippe~Baptiste, Claude~Pape, Constraint-Based Scheduling Applying Constraint Programming to Scheduling Problems, Springer New York, NY, 2012.
\newblock \href {https://doi.org/https://doi.org/10.1007/978-1-4615-1479-4} {\path{doi:https://doi.org/10.1007/978-1-4615-1479-4}}.

\bibitem{gomes2000heavy}
C.~P. Gomes, B.~Selman, N.~Crato, H.~Kautz, Heavy-tailed phenomena in satisfiability and constraint satisfaction problems, Journal of automated reasoning 24~(1) (2000) 67--100.

\bibitem{moskewicz2001chaff}
M.~W. Moskewicz, C.~F. Madigan, Y.~Zhao, L.~Zhang, S.~Malik, Chaff: Engineering an efficient {SAT} solver, in: Proceedings of the 38th annual Design Automation Conference, 2001, pp. 530--535.

\bibitem{lecoutre2007nogood}
C.~Lecoutre, L.~Sais, S.~Tabary, V.~Vidal, et~al., Nogood recording from restarts., in: IJCAI, Vol.~7, 2007, pp. 131--136.

\bibitem{10.5555/868329}
D.~Applegate, R.~Bixby, V.~Chvatal, B.~Cook, Finding cuts in the tsp (a preliminary report), Tech. rep., AT\&T Bell Laboratories (1995).

\bibitem{achterberg2005branching}
T.~Achterberg, T.~Koch, A.~Martin, Branching rules revisited, Operations Research Letters 33~(1) (2005) 42--54.

\bibitem{optalcp_params}
ScheduleOpt, \href{https://optalcp.com/docs/api/type-aliases/Parameters/}{Optalcp}, Online, optalCP's parameters (2025).
\newline\urlprefix\url{https://optalcp.com/docs/api/type-aliases/Parameters/}

\end{thebibliography}
\end{singlespace}





\end{document}